\documentclass[journal,compsoc,twoside,lettersize]{IEEEtran}

\ifCLASSOPTIONcompsoc
  \usepackage[nocompress]{cite}
\else
  \usepackage{cite}
\fi

\usepackage{graphicx}
\usepackage{amsmath}
\usepackage{mathrsfs}
\usepackage{array}
\usepackage{diagbox,xspace}

\makeatletter		% 这4行用来插入罗马数字

\newcommand{\Rmnum}[1]{\expandafter\@slowromancap\romannumeral #1@}
\makeatother

\usepackage{algorithm}
\usepackage{algorithmic}
\usepackage{booktabs}
\usepackage{makecell}
\usepackage{amssymb}
\usepackage{bm}		%% 这两个包和算法里面替换关键字为Input和Output有关
\usepackage[table]{xcolor}
\usepackage[colorlinks, citecolor=blue]{hyperref} 
\usepackage{threeparttable}
\usepackage{subfigure}

\usepackage{colortbl}
\usepackage{nicematrix}
\usepackage{tabularx}
\usepackage{multirow}
\usepackage{dsfont}
\usepackage{enumitem}
\usepackage{eucal}
\usepackage[american]{babel}
\usepackage{ragged2e}

\newcolumntype{L}[1]{>{\raggedright\arraybackslash}p{#1}}
\newcolumntype{C}[1]{>{\centering\arraybackslash}p{#1}}
\newcolumntype{R}[1]{>{\raggedleft\arraybackslash}p{#1}}

\newcommand{\vect}[1]{\mbox{\boldmath{$#1$}}}

\usepackage{amsthm}
\def\ourmethod{{DC-NeRF}\xspace}

\begin{document}
\title{
Dual-Camera All-in-Focus\\ Neural Radiance Fields}

\author{Xianrui~Luo, 
	Zijin~Wu,
    Juewen Peng,
	Huiqiang~Sun,\\  Zhiguo~Cao,~\IEEEmembership{Member,~IEEE},
	and Guosheng~Lin,~\IEEEmembership{Member,~IEEE}
% \author{Xianrui~Luo, 
% 	Zijin~Wu,
%     Juewen Peng,
% 	Huiqiang~Sun,
% 	and Zhiguo~Cao,~\IEEEmembership{Member,~IEEE}
	% and~Xin~Li,~\IEEEmembership{Fellow,~IEEE}% <-this % stops a space
\thanks{
% This work is supported by the National Natural Science Foundation of China under Grant No. . 
Corresponding author: Z. Cao.}
% \thanks{This work is supported by the National Natural Science Foundation of China under Grant No. 62106080. Corresponding author: Z. Cao.}

\thanks{X. Luo, Z. Wu, H. Sun and Z. Cao are with Key Laboratory of Image Processing and Intelligent Control, Ministry of Education; School of Artificial Intelligence and Automation, Huazhong University of Science and Technology, Wuhan, 430074, China (e-mail: \{xianruiluo, zijinwu, shq1031, zgcao\}@hust.edu.cn).}% <-this % stops a space
% \thanks{X. Li is with the Lane Department of Computer Science and Electrical Engineering, West Virginia University, Morgantown WV 26506-6109 (e-mail: xin.li@ieee.org).}
% \thanks{X. Luo, Z. Wu, J. Peng, H. Sun and Z. Cao are with Key Laboratory of Image Processing and Intelligent Control, Ministry of Education; School of Artificial Intelligence and Automation, Huazhong University of Science and Technology, Wuhan, 430074, China (e-mail: \{xianruiluo,zijinwu,juewenpeng, shq1031, zgcao\}@hust.edu.cn).}% <-this % stops a space
% % \thanks{X. Li is with the Lane Department of Computer Science and Electrical Engineering, West Virginia University, Morgantown WV 26506-6109 (e-mail: xin.li@ieee.org).}

\thanks{J. Peng and G. Lin are with Nanyang Technological University (NTU), Singapore 639798 (e-mail: \{juewen.peng, gslin\}@ntu.edu.sg).}
}
% \thanks{G. Lin is with S-Lab, Nanyang Technological University (NTU), Singapore 639798 (e-mail: gslin@ntu.edu.sg).}
% }

\markboth{Manuscript Submitted to IEEE Trans. Pattern Analysis \& Machine Intelligence; Sep~2023}%
{Shell \MakeLowercase{\textit{Luo et al.}}: Dual-Camera All-in-Focus Neural Radiance Fields}



\IEEEtitleabstractindextext{%
\begin{abstract}
\justifying
We present the first framework capable of synthesizing the all-in-focus neural radiance field (NeRF) from inputs without manual refocusing. 
Without refocusing, the camera will automatically focus on the fixed object for all views, and current NeRF methods typically using one camera fail due to the consistent defocus blur and a lack of sharp reference. 
To restore the all-in-focus NeRF, we introduce the dual-camera from smartphones, where the ultra-wide camera has a wider depth-of-field (DoF) and the main camera possesses a higher resolution. 
The dual camera pair saves the high-fidelity details from the main camera and uses the ultra-wide camera's deep DoF as reference for all-in-focus restoration.
To this end, we first implement spatial warping and color matching to align the dual camera, followed by a defocus-aware fusion module with learnable defocus parameters to predict a defocus map and fuse the aligned camera pair. We also build a multi-view dataset that includes image pairs of the main and ultra-wide cameras in a smartphone. 
Extensive experiments on this dataset verify that our solution, termed \ourmethod, can produce high-quality all-in-focus novel views and compares favorably against strong baselines quantitatively and qualitatively. 
We further show DoF applications of \ourmethod with adjustable blur intensity and focal plane, including refocusing and split diopter.
\end{abstract}

}

\maketitle

\IEEEdisplaynontitleabstractindextext

% \fi
\IEEEpeerreviewmaketitle

\IEEEraisesectionheading{\section{Introduction}\label{intro}}
\begin{figure*}
    \setlength{\abovecaptionskip}{3pt}
    \setlength{\belowcaptionskip}{5pt}
  \includegraphics[width=\textwidth]{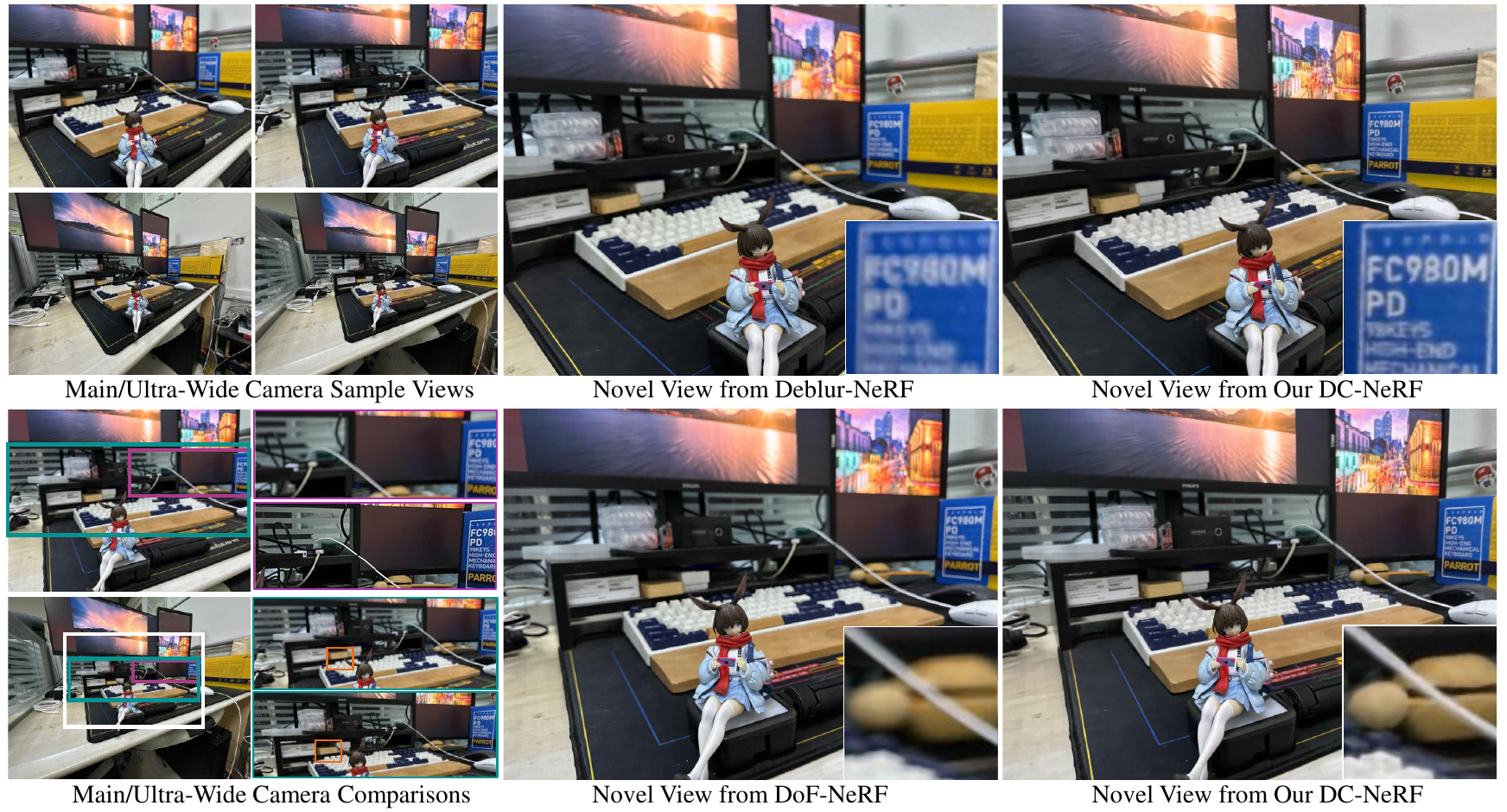}
  \caption{{We are the first to synthesize all-in-focus novel views from consistent defocus blur in smartphones.} {We show the sampled main/ultra-wide camera pair on the upper left, the first line is the main camera and the second is the ultra-wide one. We highlight the differences between the dual camera on the lower left. The two cameras exhibit significant visual differences in the field of view (indicated by the white box) and resolution. They also differ in several aspects, including variations in depth-of-field (the purple box) and the spatial and color misalignment shown in the cyan box (spatial relationship between the doll and the wooden case in the orange box).} Compared with existing methods, our \ourmethod is able to recover a sharp radiance field given a set of main camera images focused on the same target, with the help of a sub ultra-wide camera. 
  % Refer to the supplementary material for video results.
  }
  \label{fig:teaser}
\end{figure*}
\IEEEPARstart{N}{ovel} view synthesis~\cite{debevec1996modeling,gortler1996lumigraph,levoy1996light} is a popular topic commonly used in applications such as augmented reality and virtual reality. The goal is to interpolate new views from existing sparse input views. Recent methods use neural networks to learn a representation called neural radiance field (NeRF)~\cite{mildenhall2020nerf}. The network is a multilayer perceptron (MLP) that learns a function that maps spatial coordinates and view directions to volume densities and view-dependent RGB colors.
% nerf with blurry input

NeRF is typically trained on all-in-focus (AiF) images. To capture AiF images with a regular lens, one can decrease the aperture size or the focal length. Nowadays smartphones are people's go-to choice for photography, however, the focal length and the aperture size of the smartphone camera are fixed. Therefore, when the main camera and the foreground object are close, the main lens with a shallow DoF will inevitably produce defocus blur on the out-of-focus region. Furthermore, for multi-view capturing, the camera will automatically focus on the same prominent object unless manual refocusing is applied. This autofocus setting leads to the same out-of-focus object or region in all views, where the defocus blur is view-consistent.

Current AiF NeRF synthesis approaches use a deformable sparse kernel~\cite{ma2022deblur}, a Concentrate-and-Scatter technique~\cite{dofnerf} or a physical scene priors~\cite{Lee_2023_CVPR} to handle defocus blur. However, their defocus blur modeling is valid only if the camera is deliberately focused on different focal planes. The lack of sharp reference leads to their improficiencies in restoring the AiF novel views when the blur is view-consistent. 

% wzj 0901
As mentioned above, view-consistent defocus blur can often appear in images taken by smartphones, due to its autofocus setting and its large aperture and sensor that increase the blur amount. % 这里还差一句话，手机的自动对焦系统通常会固定对焦在前景上，而为了更好的图像质量而增大的传感器和光圈直径，使得view-consistent defocus blur在图像中愈发明显。
Can NeRF synthesis deal with consistent defocus blur in smartphones?
{We can circumvent the issue by introducing accessible additional inputs.}
The heterogeneous camera system on a smartphone usually contains two cameras: the wide-angle camera which is referred to as the main camera, and the ultra-wide camera. These cameras are different in their lens attributes such as the focal length, the aperture, and the field of view (FoV), 
% The discrepancies of the focal length and the aperture lead to depth-of-field (DoF) differences. 
which leads to differences in depth-of-field (DoF) and imaging quality. The main camera, which has a higher resolution sensor, captures images with higher quality than the ultra-wide camera. On the other hand, the ultra-wide camera has a lower focal length and a smaller aperture than the main camera, which in turn has a deeper DoF.  
This complementary characteristic inspires us to introduce the main/ultra-wide dual-camera system to the AiF NeRF in novel view synthesis.

% Can NeRF synthesis deal with consistent defocus blur in smartphones?
% Inspired by the DoF differences in the dual camera, we propose that the additional ultra-wide camera can extend the limited DoF of the main camera. The main camera, which has a higher resolution sensor, captures images with higher quality than the ultra-wide camera. On the other hand, the ultra-wide camera has a lower focal length and a smaller aperture than the main camera, which in turn has a deeper DoF. 
% Thus, an intuitive idea is that the ultra-wide lens can recover details from the consistent defocus blur in the main lens. 
% The superiority of the dual camera in AiF NeRF synthesis lies in 1) the two cameras work simultaneously so the cost of capturing images stays the same; % pjw: 第一点感觉有点奇怪
We propose that an additional ultra-wide camera can be of assistance to restore AiF novel views. 
The superiority of the dual camera in AiF NeRF synthesis lies in 1) the main camera and the ultra-wide camera work simultaneously and independently;
2) the ultra-wide camera with a deeper DoF provides auxiliary information for restoring the blurred regions of the main camera.
% In this work, we propose to extend the DoF of NeRF by exploiting the ultra-wide camera.
% However, as shown in Fig.~\ref{fig2}, 
However, the dual-camera system still suffers from
the following challenges: 1) the large displacement between two cameras; 2) the fusion region selection of the two cameras. 
% To process two inputs from different cameras, reference-based super-resolution methods~\cite{zheng2018crossnet,shim2020robust,lu2021masa,wang2021dual} are proposed to tackle spatial misalignment. However, they do not focus on shallow DoF scenes, which means they are not designed to deal with defocus blur. 
% Defocus deblurring methods~\cite{abuolaim2020defocus,lee2021iterative,son2021single,zamir2021restormer} % are put forward 
% aim to restore details from out-of-focus regions, but the lack of reference results in failure when the blurring amount is strong. Although current methods introduce dual-pixel image pairs, the image pairs are equivalent to stereo image pairs with a small baseline, which still provides no sharp reference for deblurring.
% Contrary to existing methods, we consider all problems above and present 

% To tackle these issues, we present \ourmethod, a feasible dual-camera framework to synthesize AiF novel views from defocus blur. 
% pjw
To tackle these issues, we present \ourmethod, a framework to synthesize AiF novel views from dual-camera image pairs.
The framework is featured by image alignment and defocus-aware fusion. 
To align differences between the dual camera, we apply spatial alignment with image registration and optical flow warping, and we also match the color of the camera pair by histogram matching. 
% To align differences between the dual camera, we apply homography warping from image registration and use optical flow warping for discrepancies by parallax. 
% To better preserve the high-quality contents of the main camera and the wide DoF region of the ultra-wide camera, we use a defocus map generated by bokeh rendering to fuse the aligned images and output AiF novel views.
To better preserve the high-quality contents of the main camera and the wide DoF region of the ultra-wide camera, defocus parameters are predicted from bokeh rendering, and a blending mask is synthesized to fuse the aligned images and output AiF novel views.
% For training, we capture three photos on each view of a scene: a main-camera photo focused on foreground, a main-camera photo focused on background, and an ultra-wide sub-camera photo captured with a small aperture and a short focal length. 
% Since the focal length and the aperture of the smartphone camera are fixed, we cannot obtain AiF ground truths for the main image directly. Therefore we implement multi-focus image fusion~\cite{qiu2019guided} to produce sharp ground truths by fusing the two main-camera photos.
As shown in Fig.~\ref{fig:teaser}, our work is the first to leverage the dual-camera module in AiF NeRF synthesis, and our method is able to recover sharp novel views when the main camera is focused on the same plane. 
% The off-the-shelf ultra-wide camera provides the defocused main photo with sharp guidance.
% To overcome the alignment issue and fuse the two cameras, we propose a viable framework to alleviate blurring artifacts. 
For training, we capture real-world data efficiently from smartphone devices. 
On each view of a scene, the main camera shoots in focal stacks, and the ultra-wide camera captures a sharp image simultaneously. 
% as shown in Fig.~\ref{fig:dataset}.
We conduct experiments to compare our solution with existing approaches. We observe that 
% current baselines are not good enough when the inputs have consistent defocus blur. % pjw
{our framework DC-NeRF is capable of harnessing ultra-wide camera to provide sharp details for consistent main-camera defocus blur in AiF NeRF synthesis and outperforms other approaches qualitatively and quantitatively, demonstrating that our framework can represent a strong baseline for dual-camera AiF NeRF synthesis.}
% our framework DC-NeRF is capable of dealing with consistent defocus blur in AiF NeRF synthesis and outperforms other approaches qualitatively and quantitatively, demonstrating that our framework can represent a strong baseline for dual-camera AiF NeRF synthesis.
% It can serve as a strong baseline for AiF NeRF synthesis from main/ultra-wide camera pair.
In summary, our contributions are as follows.
\begin{itemize}[leftmargin=*]
 % \setlength{\leftmargin}{0pt} 
    % \item[$\bullet$] To our knowledge, we are the first to define and solve the task of main/ultra-wide based AiF synthesis.
    \item[$\bullet$] To our knowledge, we are the first to explore AiF NeRF synthesis from the main and ultra-wide cameras. We propose an align-and-fuse scheme to integrate the dual-camera system into NeRF; 
    \item[$\bullet$] A strong dual-camera baseline that leverages a defocus-aware fusion module to fuse the main/ultra-wide camera pair. 
    {Our framework tackles the challenge where the main-camera inputs have consistent defocus blur and outperforms state-of-the-art approaches qualitatively and quantitatively;}
    % Our framework tackles the challenge where the inputs have consistent defocus blur and outperforms state-of-the-art approaches qualitatively and quantitatively;
    % \item[$\bullet$] We collect a dual-camera dataset for AiF novel view synthesis. For each view, we capture triplet samples consisting of the foreground-focused and background-focused main image and an ultra-wide-camera image, and we synthesize an AiF ground truth image.
    \item[$\bullet$] We collect a dual-camera dataset where each view of a scene captures triplet samples consisting of the foreground-focused and background-focused main image and an ultra-wide-camera image, and we synthesize an AiF image as ground truth.
    % \item[$\bullet$] We collect a dataset with quadruplet samples where each is composed by a pair of main-camera images that respectively focus on foreground and background, an ultra-wide-camera image, and a synthetic AiF image used as ground truth.
\end{itemize}

\section{Related Work}

% \vspace{5pt}
% \noindent\textbf{Feature Matching.}
\subsection{Novel view Synthesis}
Novel view synthesis is expected to generate new views of a scene given a set of input viewpoints. To synthesize novel views, traditional methods 
 % wzj
represent 3D geometry explicitly. % adopt explicit representations. 
Common explicit representations 
include % use 
point clouds~\cite{fan2017point,achlioptas2018learning}, voxel grids~\cite{jimenez2016unsupervised,liao2018deep,xie2019pix2vox} and meshes~\cite{kanazawa2018learning,ranjan2018generating,wang2018pixel2mesh}. % to represent 3D geometry. 
Recent approaches~\cite{NEURIPS2019_SRN, mildenhall2019local, sitzmann2019deepvoxels, 2019Neural} have shown the superiority of implicit representations in rendering photo-realistic novel views. Neural Radiance Field (NeRF)~\cite{mildenhall2020nerf} learns an implicit continuous function that models a complex 3D scene geometry and appearance. The function is represented as an MLP. The inputs of NeRF are normally images from sparse views and their corresponding poses. Follow-up studies aim to synthesize high-quality novel views from abnormal inputs. Several works~\cite{wang2021nerf,meng2021gnerf,jeong2021self} deal with the situation where camera poses are missing. The Non-linear camera model~\cite{jeong2021self} is taken into account in training NeRF. For blurry inputs, Deblur-NeRF~\cite{ma2022deblur} introduces a sparse kernel to model motion blur and defocus blur, DoF-NeRF~\cite{dofnerf} explicitly models defocus blur and enables controllable DoF effects, and DP-NeRF~\cite{Lee_2023_CVPR} adopts physical scene priors to help the {deblurring}. However, these methods require the input images to have inconsistent blur, where the images of each view are focused on different objects in different depth planes. NeRF synthesis with consistent defocus blur is yet to be explored.

% Although existing methods are effective for defocus blur, 
Compared with existing methods, we are able to synthesize AiF novel views given the view-consistent defocus blur. Current methods are under the assumption that the defocused region in one view has its corresponding sharp reference in other views, so they fail when the camera is focused on the same target. We propose to solve this problem by introducing the main/ultra-wide camera pair.
% have is effective in generating AiF results, it is time-consuming and inefficient to 
% manually change focal planes and capture required image sequences using the same smartphone lens. % is time-consuming and inefficient. 
% Therefore, we propose a user-friendly AiF synthesis routine to produce an AiF image from a main and ultra-wide image pair that can be captured at the same time.
\subsection{Bokeh Rendering}
% Defocus % Bokeh 
Bokeh rendering has seen progress in recent years. Early methods apply ray tracing~\cite{pharr2016physically,yang2016virtual} which models the physical ray integration in cameras. However, they have large computation costs. Recently a large dataset EBB!~\cite{ignatov2020rendering} is proposed for training. Several works train a neural network and directly output the shallow DoF effects, using GAN~\cite{qian2020bggan} or intermediate defocus estimation~\cite{srinivasan2018aperture,luo2023defocus} to assist the prediction.
To achieve controllable shallow DoF, depth maps~\cite{wadhwa2018synthetic,xiao2018deepfocus,wang2018deeplens,peng2021interactive,xian2021ranking,dutta2021depth,peng2022bokehme} are introduced to adjust the defocused % bokeh 
intensity and the focused object with the input of aperture size and focal distance.  
Although disparity maps are easy to obtain with the progress in single image depth estimation~\cite{ranftl2020towards}, the depth information of the background objects is lacking due to occlusion from a single view. To address the problem, Multiplane Image is implemented to render the result with image inpainting. To fully exploit 3D geometry, several methods propose to synthesize shallow DoF on NeRF model~\cite{mildenhall2021nerf,wang2022nerfocus,dofnerf}. 

In this work, our defocus-aware fusion module is inspired by classical rendering. With the off-the-shelf depth map from NeRF, we predict a defocus map from the learnable rendering parameters. 

\subsection{Defocus Deblurring}
% need rewrite
Traditional defocus deblurring methods harness image priors to estimate a defocus map~\cite{shi2015just,park2017unified,kim2017dynamic}, then the defocus map is used as guidance to deblur the image from non-blind deconvolution~\cite{levin2007image,dong2021dwdn}. 
Recent methods deblur an image directly by neural networks. To train the network, a real-world dual-pixel dataset~\cite{abuolaim2020defocus} is introduced. Iterative adaptive convolutions~\cite{lee2021iterative} and kernel-sharing parallel atrous~\cite{son2021single} convolutions are proposed to improve the deblurring performance. A novel multi-scale feature extraction model~\cite{zamir2022learning} is proposed to learn enriched contextual information. Video sequences~\cite{abuolaim2021learning} and depth images~\cite{pan2021dual} are introduced as guidance for defocus deblurring. The transformer architecture is applied~\cite{zamir2022restormer} and achieves the state of the art in defocus deblurring.
For single-view defocus deblurring, EasyAIF~\cite{luo2022point} and DC2~\cite{alzayer2023defocuscontrol} utilize the ultra-wide camera to restore the large defocus blur of the main camera. However, both methods collect a prerequisite dataset for training. 
% For the multi-view novel view synthesis, this training results in an extra pre-training step. 
{Both methods are meant to solve single-view dual camera defocus deblurring, therefore, they need an extra single-view dataset for pre-training. Then they apply the pre-trained model to deblur the inputs, the deblurred results then are used for NeRF training. This is an extra step in need of an extra dataset, increasing the total training time and cost.}
% The dual camera system is used in AiF image restoration~\cite{luo2022point} to solve the single-view scenario. 

Current defocus deblurring methods do not take multi-view 3D geometry into account. 
To our knowledge, we are the first to integrate the main/ultra-wide dual-camera setup into multi-view AiF NeRF synthesis, {and our method is able to recover from consistent main-camera defocus blur by utilizing ultra-wide inputs.}
% To our knowledge, we are the first to integrate the main/ultra-wide dual-camera setup into multi-view AiF NeRF synthesis, and our method is able to recover from consistent defocus blur.

% Although defocus deblurring can restore a sharp image, it requires a large dataset for training.

% it only utilizes a single image or a dual-pixel image pair instead of a sharp reference image, which is not sufficient to recover details from a large blur. The deblurring model also requires training on a large dataset. Furthermore, these methods do not take 3D geometry into account. The dual-pixel image pairs are equivalent to stereo image pairs with a small baseline, while NeRF uses the multi-view input with a large baseline. 

% wzj: 前后文不提bokeh的话这里是不是用defocus更好？
% \subsection{Defocus Rendering} % \subsection{Bokeh Rendering}

% \subsection{Reference-based Application}
% Using a reference image to boost performance has been explored in various low-level tasks. Reference-based image super resolution use images from cameras on different devices~\cite{shim2020robust,lu2021masa}, and the reference image is used as guidance in image inpainting~\cite{zhou2021transfill}. The main/telephoto dual camera pair is used in image super resolution~\cite{wang2021dual,zhang2022self} to take advantage of the different camera resolutions, and it is also applied in bokeh rendering~\cite{luo2020wavelet}. 

% \section{Background, Motivation, and Overview} \label{sec:background and motivation}
 \section{Preliminary}
We follow the basic principal of NeRF ~\cite{mildenhall2020nerf} to represent the 3D scene using a continuous function $R_{\Theta}$. Parameterized by an MLP, $R_{\Theta}$ maps a 3D position $\vect{x}$ and a 2D view direction $\vect{d}$ to color $\vect{c}$ and volume density $\sigma$, 
\begin{equation}
    \begin{aligned}
     (\vect{c},\sigma)= R_{\Theta}(\vect{x},\vect{d})\,.
    \end{aligned}
    \label{nerf}
\end{equation}

To query the RGB color of a pixel, NeRF first specifies a ray $\vect{r}(t)=\vect{o}+t\vect{d}$ from camera projection center $\vect{o}$ along the viewing direction $d$. The color value of ray $\vect{r}(t)$ can be represented as an integral of all colors along the ray following the classical volume rendering technique~\cite{kajiya1984ray}. It can also be approximated as the sum of weighted radiance at $k$ points $\left \{ t_i \right \}_{i=1}^k$ on ray $\vect{r}(t)$:
\begin{equation}
    \begin{aligned}
        % \hat{C}(\vect{r})=\sum_{1}^{k}T(t_i)(1-\exp(-\sigma(t_{i+1}-t_i)))\vect{c}(\vect{r}(t_i),\vect{d})\,, 
        % wzj0503
        % \hat{C}(\vect{r})=\sum_{1}^{k}T(t_i)(1-\exp(-\sigma\delta_i))\vect{c}(\vect{r}(t_i),\vect{d})\,, 
        \hat{C}(\vect{r})=\sum_{i=1}^{k}T_i(1-\exp(-\sigma\delta_i))\vect{c}(\vect{r}(t_i),\vect{d})\,, 
    \end{aligned}
\end{equation}
where $\delta_i=t_{i+1}-t_i$ is the distance between the two sampled points. 
$T_i=\exp(-\sum_{j=1}^{i-1}\sigma \delta_j)$ denotes the accumulated transmittance along the ray. 
% $T_i=\exp(-\prod_{j=1}^{i-1}\sigma \delta_j)$ denotes the accumulated transmittance along the ray. 
% where $T(t_i)=\exp(-\prod_{j=1}^{i-1}\sigma(\vect{r}(t_j),t_{j+1}-t_j))$ denotes the accumulated transmittance along the ray. 

Note that the volume rendering technique is fully differentiable, NeRF adopts L2 loss between the rendered color $\hat{C}_{\vect{r}}$ and the ground-truth color $C_{\vect{r}}$ corresponding to the pixel to optimize the MLP, 
\begin{equation}
    \begin{aligned}
    L=\sum_{r=1}^{N}\Vert C(\vect{r})-\hat{C}(\vect{r})\Vert_2^2\,.
    \end{aligned}
\end{equation}

The vanilla NeRF fails to represent the 3D scene when the input views are blurred. It assumes a pinhole camera model, where the color of a pixel $\hat{C}(\vect{r})$ is calculated by points on a single ray $\vect r(t)$, so it does not model the defocus blur. 
% However, to model the defocus blur, the correlation between the specified pixel and its surrounding rays should be considered.  

%%%%%%%%%%%%%%%%%%%%%%%%%%%%%%%%%%%%%%%%%%%%%%%%%%%%%%

% \section{From Barycentric to Probabilistic Coordinate Fields} \label{sec: From Barycentric to Probabilistic Coordinate Fields}

\section{Method}
\begin{figure*}
  \centering
  \setlength{\abovecaptionskip}{3pt}
  \setlength{\belowcaptionskip}{0pt}
  \includegraphics[width=\linewidth]{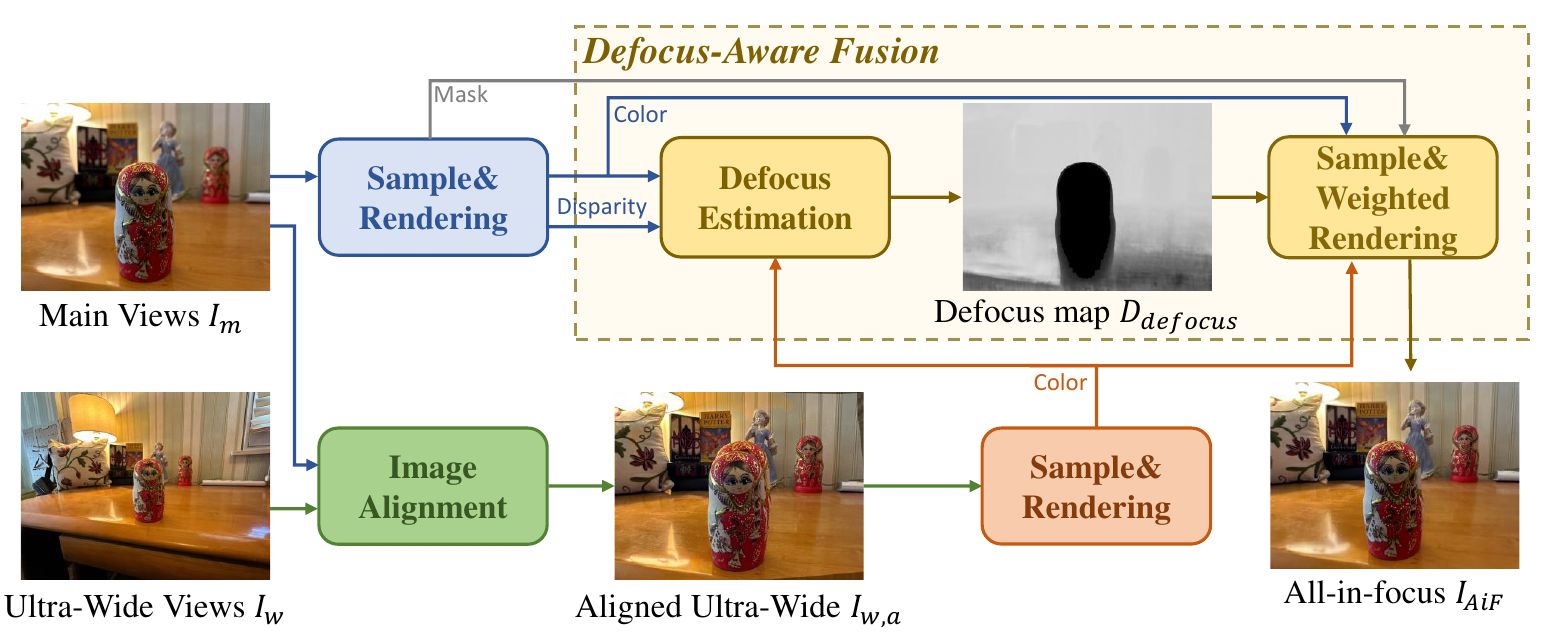}
  \caption{
  {Our framework includes two modules: image alignment, and defocus-aware fusion.}
  Homography and flow warping are implemented to align the main/ultra-wide image pair $I_m$ and $I_w$. Then we adjust the color of the warped ultra-wide image from histogram matching. To fuse the high-quality shot of the main camera view and the deep DoF information of the ultra-wide view, we propose a defocus-aware fusion network to estimate defocus parameters from bokeh rendering. The network then predicts a blending mask fuses the dual-camera view in volume rendering to generate the AiF novel views $I_{AiF}$.}
  \label{fig_pipeline}
\end{figure*}
\begin{figure}
  \centering
  \setlength{\abovecaptionskip}{3pt}
    \setlength{\belowcaptionskip}{0pt}
  \includegraphics[width=\linewidth]{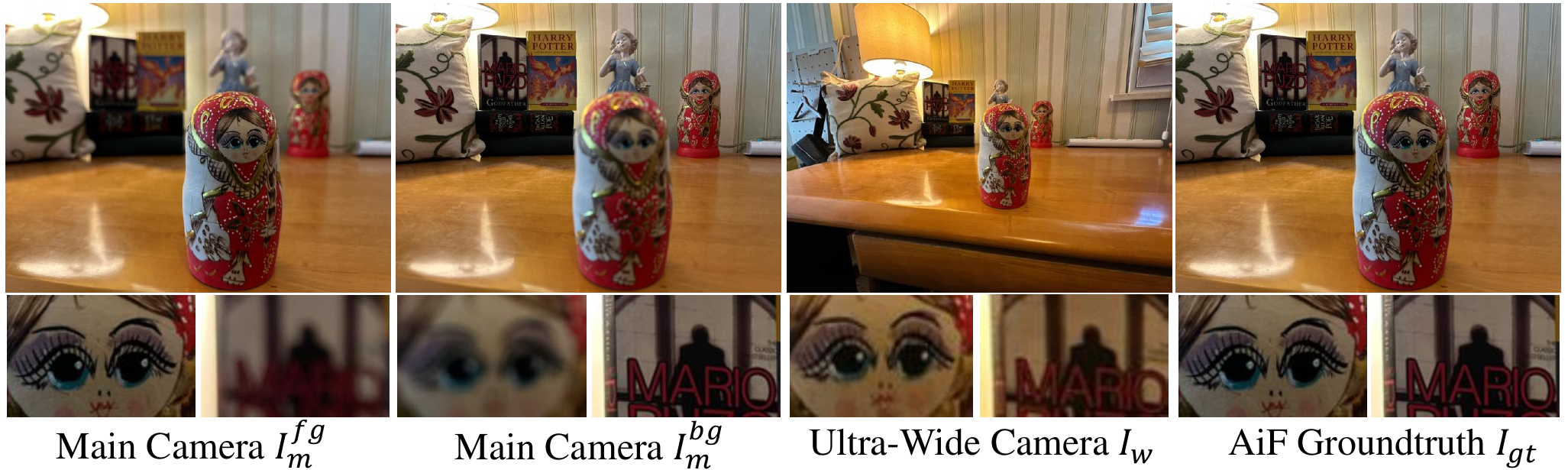}
    \caption{{An example view of the dataset.} For each view, we capture a foreground-focused main image $I_{m}^{fg}$, and a background-focused $I_{m}^{bg}$. We simultaneously capture the ultra-wide image $I_w$ with deep DoF. The main camera focal stack is used to synthesize the AiF ground truth $I_{gt}$.}
  % Since the focus breathing effect causes wider FoV in $I_{m}^{bg}$, we align $I_{m}^{bg}$ to $I_{m}^{fg}$ using flow warping, and the two images are fused to synthesize the AiF ground truth image $I_{gt}$.}
  \label{fig:dataset}
\end{figure}
\begin{figure}
  \centering
  \setlength{\abovecaptionskip}{3pt}
    \setlength{\belowcaptionskip}{0pt}
  \includegraphics[width=\linewidth]{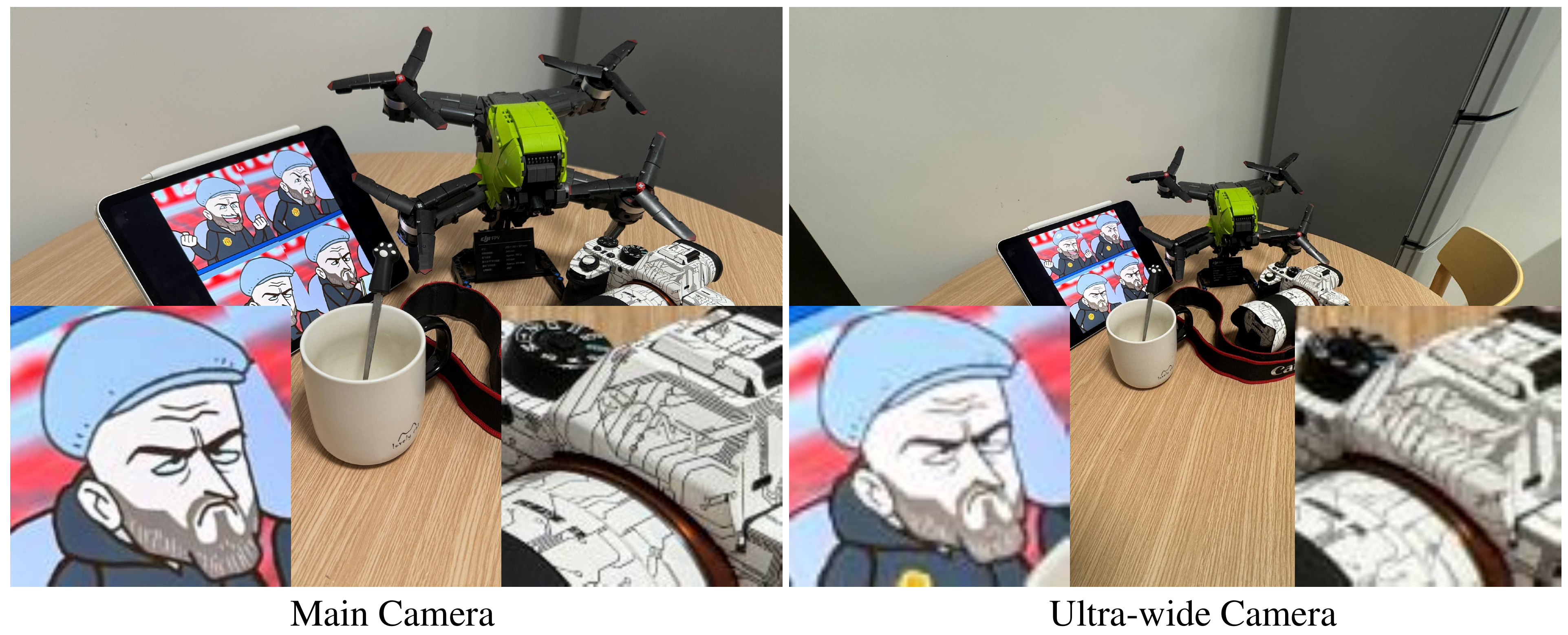}
    \caption{{The advantage of the main camera on the ultra-wide camera. Although the ultra-wide camera has a larger DoF than the main camera, we still prefer the main camera for the focused area, because the main camera captures finer details due to sensor resolution.}}
  % Since the focus breathing effect causes wider FoV in $I_{m}^{bg}$, we align $I_{m}^{bg}$ to $I_{m}^{fg}$ using flow warping, and the two images are fused to synthesize the AiF ground truth image $I_{gt}$.}
  \label{fig:main_advantage}
\end{figure}
% Our goal is to extend the NeRF's DoF from consistent smartphone blur inputs. As mentioned, the NeRF model should be able to predict the in-focus region in the camera and extend its DoF by transferring sharp details from other views.
% Our goal is to extend the NeRF's DoF from defocus blur inputs. As mentioned, the model is expected to predict the in-focus region of each view and extend its DoF by transferring sharp details from other views.
% However, to train such a model, we are faced with a problem: if the focused object is the same for all views, to acquire sharp contents for extending DoF, we require a camera to capture images with different apertures or focal lengths, which is not available on smartphones. 
% The lens on a smartphone has a fixed aperture and focal length. 

% To resolve this issue, we introduce the ultra-wide camera to extend the DoF of NeRF. The dual-camera framework is able to restore the sharp contents in the shallow DoF main camera and keep the high-quality in-focus region intact.
% We first collect a dataset of the main/ultra-wide camera pair. We utilize the two cameras simultaneously, which used the ultra-wide image as guidance for deep DoF efficiently.
As shown in Fig.~\ref{fig_pipeline}, our framework \ourmethod consists of two components: image alignment, and defocus-aware fusion. 
For each view, we align the ultra-wide image $I_w$ to the main camera image $I_m$. 
% Image registration and optical flow are applied to solve the spatial misalignment, and color matching is implemented. Homography warping from registration focuses on a global scale and produces a coarse result, and flow warping is stable under parallax issues brought by different depth planes. 
To address the spatial misalignment between the main and ultra-wide cameras, we utilize both homography and optical flow. A global homography is estimated to coarsely warp the ultra-wide image, then optical flow provides a pixel-level alignment robustness to parallax from objects at different depths.
% Additionally, color matching is applied to adapt the color profiles of the two camera inputs. 
% To solve the misalignment between the main camera $I_m$ and the ultra-wide camera $I_w$, we apply both homography warping and flow warping to output the aligned ultra-wide $I_{w,f}$. Homography warping focuses on a global scale and produces a coarse result, and flow warping is stable under parallax issues brought by different depth planes. 
To fuse the high-quality in-focus region of the main camera view and the extended DoF details from the ultra-wide camera view, we 
% introduce defocus estimation
propose a defocus-aware fusion network
to predict the defocus parameters along with a blending mask. The details are described in the following.
% The defocus map is learned only from the RGB images of the main/ultra-wide camera pair as input.

\subsection{Dataset Collection} 
\label{sec:dataset}
To train our NeRF model \ourmethod, we capture the main/ultra-wide image pair and synthesize the ground truth AiF image for evaluation.
We use iPhone 14 Pro as our camera platform and capture a dataset of 7 scenes. 
For each view of a scene, we choose two focus planes for the main camera and shoot two images that form a focal stack. We simultaneously capture the ultra-wide image $I_w$ with a smaller aperture and a larger DoF. 
With three images per view, aligning the two main images is necessary to account for the phenomenon of focus breathing, where the FoV narrows when shifting focus to a closer object. Therefore, optical flow~\cite{teed2020raft} is used to align $I_{m}^{bg}$ to $I_{m}^{fg}$. Finally, the two main images serve as inputs to a multi-focus image fusion method~\cite{qiu2019guided} and output a fusion mask $M_{fuse}$ to synthesize the AiF ground truth $I_{gt}$ for comparison:
\begin{equation}
    \begin{aligned}
    I_{gt}=M_{fuse}\cdot I_{m}^{fg}+(1-M_{fuse})\cdot I_{m}^{bg}\,,
    \end{aligned}
\end{equation}
where $I_{m}^{fg}$ and $I_{m}^{bg}$ are the main images focused on foreground and background, respectively.
Noted that we shoot the scene from the main camera twice for dataset collection. In NeRF training, we only need to capture the scene once for each view, where we acquire one main image and one ultra-wide image.
In all, we collect a dataset with $7$ scenes. For each view, a image set consists of $I_{m}^{fg}$, $I_{m}^{bg}$, $I_w$, and $I_{gt}$, as shown in Fig.~\ref{fig:dataset}. {Although the ultra-wide camera itself provides with large DoF, we prefer to use the focused area in the main camera, because the main camera captures finer quality details, as shown in Fig.~\ref{fig:main_advantage}}

\subsection{Image Alignment}
\begin{figure}[htp]
  \centering
  \setlength{\abovecaptionskip}{3pt}
    \setlength{\belowcaptionskip}{0pt}
  \includegraphics[width=\linewidth]{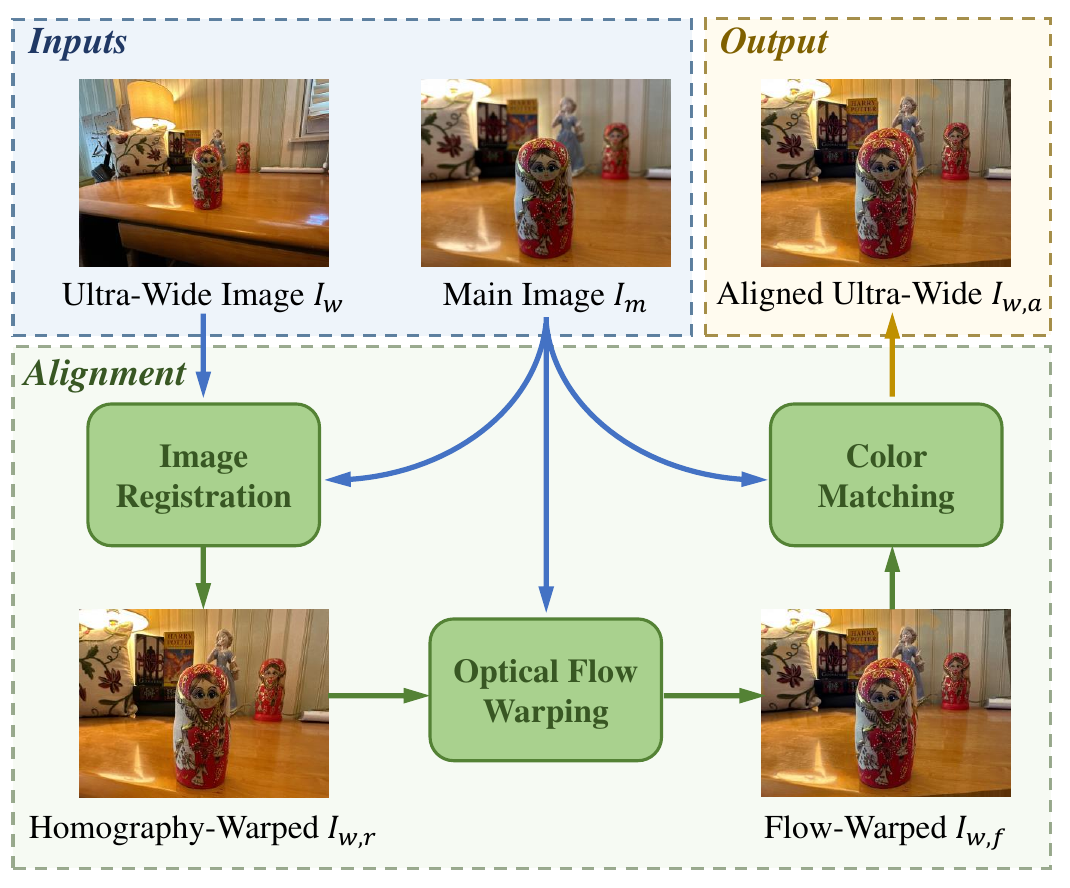}
  \caption{{The alignment between an ultra-wide image view $I_w$ and a main image view $I_m$.} We use image registration to warp the ultra-wide $I_w$ from the image level, and the optical flow is applied to make up for parallax from different depth planes. Histogram matching is then used to match the color of the main view and the spatially warped ultra-wide view.}
  \label{spatial}
\end{figure}

The main camera and the ultra-wide camera have large gaps, as shown in Fig.~\ref{spatial}. To align the two images spatially, an intuitive solution is to use image registration. The goal of image registration is to find robust correspondences between the two images. 
We use SIFT~\cite{lowe1999object} to find keypoints with distinctive features, then we establish~\cite{mishchuk2017working} {correspondences} and filter~\cite{zhao2019nm} {them} between the keypoints. 
% We use SIFT~\cite{lowe1999object} to find keypoints with distinctive features, then we establish~\cite{mishchuk2017working} and filter~\cite{zhao2019nm} correspondences between the keypoints. 
% We use SIFT~\cite{lowe1999object} to find keypoints with distinctive features, and pre-correspondences are established between $I_w$ and $I_m$ from HardNet~\cite{mishchuk2017working} descriptors. Then NM-Net~\cite{zhao2019nm} is implemented to filter the outlier correspondences. 
With the filtered correspondences, we calculate a homography matrix using RANSAC~\cite{fischler1981random}, and the matrix warps $I_w$ to $I_m$.
Image registration uses keypoints correspondences to achieve an image-level alignment by homography. 
However, it is impractical to assume the objects in the scene are of the same depth plane for shallow DoF inputs. 
The parallax from different depth planes is not taken into account by image-level homography.
% The parallax of an object increases from foreground to background, and this disparity from different depth planes is not taken into account by image-level homography. 

% Homography warping aligns the input at the image-level, which can restore image-level attributes such as the FoV. However, in our smartphone AiF synthesis task, prominent foreground/background relationships result in a larger disparity on background regions than on foreground ones. Therefore it is impossible to align the main/ultra-wide image pair $I_m$ and $I_w$ using only a single homography warping. 
To align objects at varying depths, we utilize optical flow for its robustness to parallax differences. Specifically, we apply the RAFT~\cite{teed2020raft} model to estimate the optical flow. As shown in Fig.~\ref{spatial}, homography provides coarse alignment of image-level attributes like FoV, and the flow warping aligns objects across different depth planes.

% Homography warping provides a coarse image-level alignment for our pipeline. 
% as a single homography alignment is based on the assumption that every object is on the same depth plane. 
% The warping results $I_{w,r}$ and $I_m$ still have unaligned regions due to large parallax. 
% Therefore, we apply a pixel-wise warping alignment with an optical flow field.
% Pixel-wise warping is insensitive to varying parallax from different depth planes. 

\noindent\textbf{Occlusion Mask.}
Although optical flow is effective to establish pixel-level alignment, it performs poorly when the pixels in one image are occluded. We use forward-backward consistency check~\cite{chen2016full} to alleviate this problem. The idea is that we warp a point from the source image to the target image using forward flow, and the point is warped back by the backward flow. The confidence is high if the distance between the reprojected point and the original point is smaller than a threshold. We compute the flow $F_{I_{w,r} \rightarrow I_m}$ from the ultra-wide image to the main image as the forward flow, $F_{I_m \rightarrow I_{w,r}}$ as the backward flow, and estimate a confidence map:
\begin{equation}
    \begin{aligned}
     M_c = t\cdot \left \| f_w(F_{I_m \rightarrow I_{w,r}},f_w(F_{I_{w,r} \rightarrow I_m},p)-p \right \|_2\leq 1.0\,,
    \end{aligned}
    \label{consistency}
\end{equation}
where $p$ is a point in $I_{w,r}$, $f_w$ is the warping function, and $t$ is a scaling threshold to control how much inconsistency is tolerated. 

\noindent\textbf{Color Alignment.}
% Although we fix the exposure and the white balance of the two cameras, the main images and the ultra-wide images still have color discrepancies due to different sensor features. 
The main image and the ultra-wide image have color discrepancies due to different sensor features. 
To mitigate the color mismatch, We align the color of the ultra-wide image $I_w$ to the main image $I_m$ by matching the image histograms. Histogram matching is to match the histogram's cumulative distribution function. For an input RGB image, the cumulative distribution function can be acquired by:
\begin{equation}
    \begin{aligned}
     S(v_i)=\sum_{j=0}^i h(v_j) \,.
    \end{aligned}
\end{equation}
where $v_i$ is an RGB color, $n_i$ is the frequency of the color, $h(v_i)=\frac{n_i}{n}$ is the histogram. To match the distribution function of the ultra-wide $S_w$ to the main $S_m$, for each level $i\in[0,255]$, we find a corresponding $k$ that fits $S_w(v_i)=S_m(v_k)$.

\subsection{Defocus-Aware Fusion}
\label{sec:defocus}
\begin{figure}[htp]
  \centering
  \setlength{\abovecaptionskip}{3pt}
    \setlength{\belowcaptionskip}{0pt}
  \includegraphics[width=\linewidth]{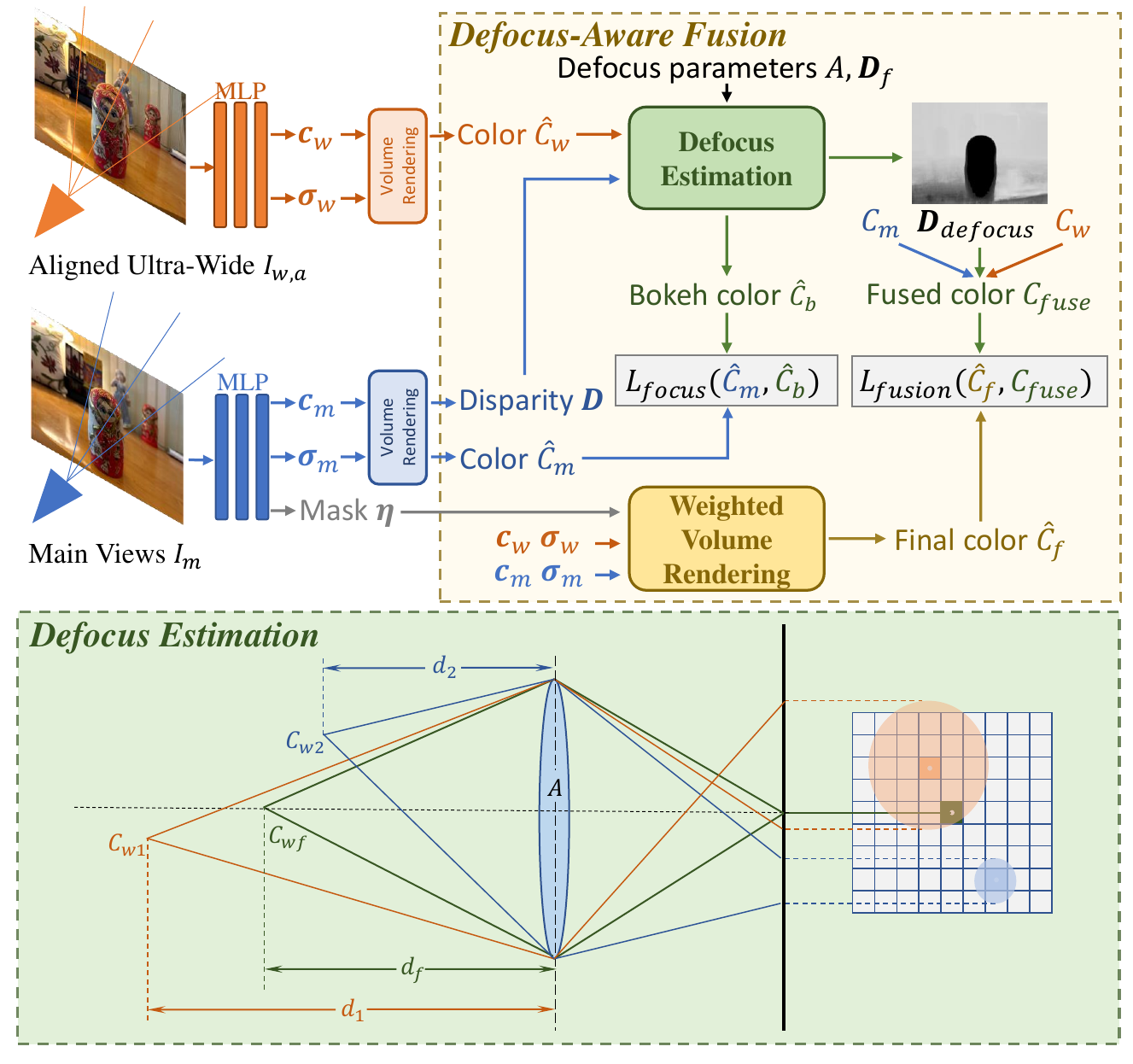}
  \caption{{The defocus-aware fusion module.} 
  % We use two MLPs for two cameras. 
  The ultra-wide color $C_w$ and the defocus parameters are used as inputs to render bokeh as Eq.~\ref{eq:scatter}. The rendered color $C_b$ is supervised by the main camera color $C_m$. To calculate the defocus map $D_{defocus}$, the main-camera radiance field also renders a disparity map $\vect{D}$. 
  The blending weight field outputs a blending mask $\eta$ to render the final color by weighted volume rendering.
  % To achieve fusion of the novel view, the blending weight field outputs a blending mask $\eta$ for weighted volume rendering.
  $D_{defocus}$ is only for training supervision.
  }
  \label{fusion}
\end{figure}

Now that we align the dual camera, the camera pair is then used to synthesize AiF novel views. We do not have an AiF ground truth in NeRF training, so we address the problem with a defocus-aware fusion module, predicting a blending mask to get the best of both cameras, as shown in Fig.~\ref{fusion}.
% Now that we align the two cameras, we need to fuse the sharp contents in the camera pair. 
% Specifically, we need to transfer the extended DoF region from the ultra-wide image and save the already sharp in-focus region from the main image. 

To fuse the aligned dual camera pair, we transfer the extended DoF region from the ultra-wide image and save the already sharp in-focus region from the main image. 
In that regard, we propose to predict the degree of defocus blur of each pixel from the main image, which is also referred to as the defocus map of the main image. 

First we introduce bokeh rendering. Bokeh is the aesthetic quality of the blur in the out-of-focus region, caused by Circle of Confusion (CoC). Under the assumption of a thin lens, the blur radius of a pixel's CoC is calculated by
\begin{equation}
    \begin{aligned}
    % r =  A\frac{f\left|D_{depth}-D_{focus}\right|}{D_{depth}(D_{focus}-f)}\,.
    r=af|\frac{1}{D_{focus}}-\frac{1}{D_{depth}}|=af|D_f-D|
    \,,
    \label{eq:radius}
    \end{aligned}
\end{equation}
where $a$ is the camera aperture radius, $f$ is the focal length, $D_{depth}$ is the depth of the pixel, $D_{focus}$ is the depth of the in-focus region. We use the disparity map to replace the depth map, where $D_f$ is the focused disparity, and $D$ is the disparity of the pixel. The illustration of the CoC generation in a thin lens model is shown in Fig.~\ref{fusion}.
The focal length of a camera is fixed, so the variables for the blur radius are the aperture size $a$ and the focal distance $D_f$. $a$ controls the intensity of the blur, and $D_f$ is the disparity of the focused object. 

Now that we calculate the blur radius of each pixel, bokeh rendering is implemented by the pixel-wise $scatter$ operation~\cite{peng2022bokehme}.
Classical bokeh rendering can be categorized into $gather$ and $scatter$ operations. We choose $scatter$ because it is closer to the physical behavior of light. We first transform the RGB value into linear radiance from gamma correction $I_g=I^\gamma$, where $\gamma$ is the intensity of the exponential function. For each pixel $x$ of the linear radiance $I_g$, we set the light it emits as a weighted kernel $K_x$ with the pixel $x$ in its center. For each element in $K_x$, we define its offset to the point $x$ as $\epsilon$, 
% and the weight $k_{\epsilon}$ is defined as:
and the weight is defined as:
\begin{equation}
    \begin{aligned}
    K_{x}(\epsilon)=\frac{H(r_x-D_{\epsilon})}{{r_x}^2} 
    \,,
    % K_{x}(\epsilon)=k_{\epsilon}=\frac{H(r_x-d_{\epsilon})}{{r_x}^2} 
    % \,,
    \label{eq:kernel}
    \end{aligned}
\end{equation}
where $r_x$ is the blur radius of the pixel, $D_{\epsilon}$ is the distance between the center pixel and the pixel element in the kernel. $H(x)$ is a function where $H(x)=0$ if $x<0$, and $H(x)=1$ if $x\geq0$. To achieve a smooth weight distribution, we replace $H(x)$ with $H(x)=\frac{1}{2}+\frac{1}{2}\tanh(\beta\cdot x)$.
The size and the weights of the kernel $K_x$ are different according to the pixel $x$. 
Now we calculate the bokeh of the pixel $x$: 
\begin{equation}
    \begin{aligned}
    C_b(x)=(\sum_{\epsilon}I_g(x+\epsilon )K_{x+\epsilon }(-\epsilon ))^\frac{1}{\gamma}
    \,,
    \label{eq:scatter}
    \end{aligned}
\end{equation}
where $\epsilon$ values are the offsets of the pixel $x$, and $x+\epsilon$ are the pixels which scatter their radiance to the pixel $x$. 
% In other words, $k_{x(x+\epsilon)}>0$.
The core idea of bokeh rendering is to reassign each pixel to its neighboring region where the distance between the two pixels is less than the blur radius of the specified pixel. For a pixel, its RGB value is determined by a weighted rendering of surrounding radiated pixels, and the weights are calculated by whether the pixel is within the blur radius of the nearby pixels.

Now that we are familiar with bokeh rendering, we integrate the rendering process into defocus map estimation. 
The defocus map represents the blur intensity of each pixel, which is calculated by 
\begin{equation}
    \begin{aligned}
		D_{defocus} =  A|D - D_f|\,,
  \label{eq:defocus}
  \end{aligned}
\end{equation}
where $D_f$ is the focal distance, $D$ is the disparity or inverse depth of a pixel, and $A=af$ controls the blur intensity. 
% 要对应把图改了
The disparity map can be directly acquired from the NeRF model. For $k$ points $\left \{ t_i \right \}_{i=1}^k$ on ray $\vect{r}(t)$,
\begin{equation}
		D = \frac{1}{\sum_{i=1}^{k}T_i(1-\exp(-\sigma\delta_i))t_i} \,,
		% D = \frac{\sum_{i=1}^{k}T_i(1-\exp(-\sigma\delta_i))}{\sum_{i=1}^{k}T_i(1-\exp(-\sigma\delta_i))t_i} \,,\\
  \label{eq:disparity}
\end{equation}
where $\delta_i=t_{i+1}-t_i$ is the distance between the two sampled points. $T_i=\exp(-\prod_{j=1}^{i-1}\sigma \delta_j)$ denotes the accumulated transmittance along the ray.
% To estimate the defocus map, we propose to learn the blur intensity parameter $A$ and the focal distance $D_f$.
% To learn these parameters, for an input ray, we sample its surrounding area and $scatter$ its radiance to implement bokeh rendering. 
% To learn these parameters, for an input ray, we sample its surrounding area and implement bokeh rendering as mentioned. 
To learn the defocus parameters $A$ and $D_f$, we use the ultra-wide ray and the parameters as inputs, sampling its surrounding area and implementing bokeh rendering as mentioned. 
We use the shallow DoF rays from the main camera as supervision for the defocus parameters estimation. 

\begin{figure}[htp]
  \centering
  \setlength{\abovecaptionskip}{3pt}
    \setlength{\belowcaptionskip}{0pt}
  \includegraphics[width=\linewidth]{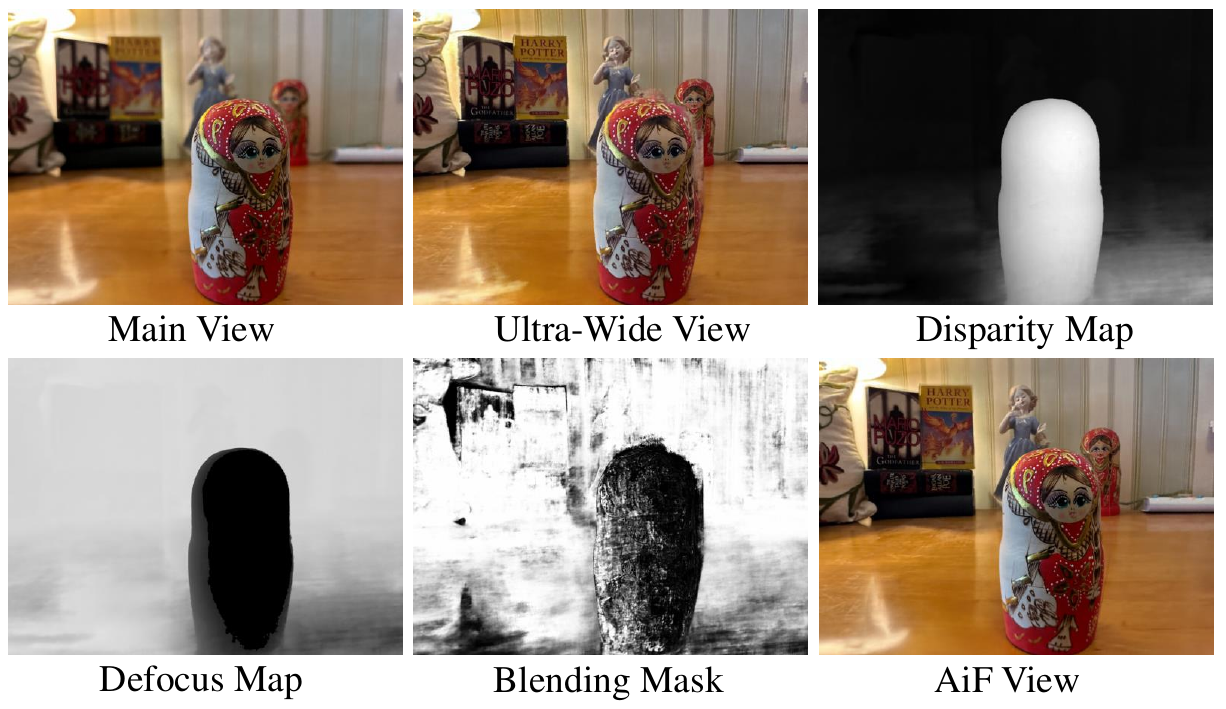}
  \caption{{The visualization of the defocus-aware fusion.}
  The defocus map is calculated by the defocus parameters and the disparity map.
  % The fusion module predicts the defocus parameters, and the parameters are used with the predicted disparity map to calculate a defocus map for blending. 
  The blending weight field outputs the blending mask and generates AiF novel views from weighted volume rendering. Note that the defocus map is for training, we only predict the blending weight for novel views, and the blending mask here is only for visualization because the weight is integrated implicitly in weighted volume rendering.}
  \label{blendmask}
\end{figure}

% 不知道要不要写
% For implementation, we first train a NeRF model with the wide DoF ultra-wide camera rays as supervision, then we implement bokeh rendering on this model with the shallow DoF main camera rays as supervision. 
After we acquire a defocus map, the defocus information can be harnessed for fusing the main rays and ultra-wide rays. Since we aim to keep the sharp object in the main image, the expected defocus value of the sharp object should be zero, which means the object is not blurred. 
% We show the results in Fig.~\ref{blendmask}.
We use the defocus information to supervise the novel-view fusion mask prediction. A blending weight field is proposed to output the weight $\eta$ to fuse the main camera scene and the ultra-wide scene. $\eta$ assigns a low weight to the blurred region, which means the ultra-wide scene gets a {high} weight. The rendering with the blending weight is defined as:
\begin{equation}
    \begin{aligned}
    \hat{C_f}(\vect{r})=\sum_{1}^{k}T_f(t_i)volume_f\,,
    % \hat{C_f}(\vect{r})=\sum_{1}^{k}T_f(t_i)\alpha_f\vect{c_f}(\vect{r}(t_i),\vect{d})\,,
    \label{eq:blend}
    \end{aligned}
\end{equation}
where 
$T_f(t_i)=\exp(-\sum_{j=1}^{i-1}\sigma_m \sigma_w\delta_j)$,
% $T_f(t_i)=\exp(-\prod_{j=1}^{i-1}\sigma_m \sigma_w\delta_j)$,
and
$volume_f$ is the fusion of RGB color from the main and ultra-wide scene with $\eta(t_i)$ as weights:
\begin{equation}
    \begin{aligned}
    % \sigma_f\vect{c_f}(\vect{r}(t_i),\vect{d})&=\eta(t_i)(1-\exp(-\sigma_w(t_{i+1}-t_i)))\vect{c_w}(\vect{r}(t_i),\vect{d}) \\
    % &+(1-\eta(t_i))(1-\exp(-\sigma_m(t_{i+1}-t_i)))\vect{c_m}(\vect{r}(t_i),\vect{d})  
    volume_f&=\eta(t_i)(1-\exp(-\sigma_w\delta_i))\vect{c_w}(\vect{r}(t_i),\vect{d}) \\
    &+(1-\eta(t_i))(1-\exp(-\sigma_m\delta_ i)\vect{c_m}(\vect{r}(t_i),\vect{d})  
    \,.
    \label{eq:weight}
    \end{aligned}
\end{equation}
The weight $\eta(t_i)$ is the blending mask corresponding to the ray $\vect{r}(t_i)$. 
{$\eta(t_i)$ is  predicted from an MLP. For each input ray, we predict a blending weight, the MLP also predicts the rendered radiance, and then the weight along with the radiance is used to render the final color. }
We demonstrate the mask in Fig.~\ref{blendmask}.

\subsection{Training Details}
% need revision
\noindent\textbf{Optimization.}
In the first stage, we train two NeRF models. One model is supervised by the rays from the aligned ultra-wide image $I_{w,a}$, and the other model is supervised by the main camera rays. Both models use the regular MSE loss:
\begin{equation}
    \begin{aligned}
    \mathcal L_{recon} =\sum_{r=1}^{N}\Vert C(\vect{r})-\hat{C}(\vect{r})\Vert_2^2\,,
    \end{aligned}
\end{equation}
For the training of the ultra-wide NeRF model, we use the confidence map $M_c$ mentioned above to mitigate the influence of wrongly aligned regions. We discard the rays from the low-confidence region from the start of training.
% For rays from the main image $I_m$, we use the following loss:

% where we use depth loss to regulate the generated disparity map, as many previous works have done~\cite{}. We use the disparity map from Midas~\cite{} to supervise the training.
{The second stage is to learn the defocus parameters. The goal is to predict the amount of defocus blur on each main-camera ray. The defocus blur model is described in Sec \ref{sec:defocus}. The parameters of the two NeRF models are fixed. The amount of blur provides guidance where main-camera rays are sharp, and we aim to replace the rest with aligned ultra-wide rays.}
For the learnable focus settings, we use the loss function
\begin{equation}
    \begin{aligned}
    \mathcal{L}_{\mathit{focus}} =\mathcal{L}_{ssim}(\hat C_m(\vect{r}),\hat C_b(\vect{r}))\,,
    \end{aligned}
\end{equation}
where $L_{ssim}$ is the structural similarity (SSIM) loss~\cite{DBLP:journals/tci/ZhaoGFK17}, {$\hat C_b(\vect{r})$ is the rendered bokeh from the rendered ultra-wide camera ray $\hat C_w(\vect{r})$} and $\hat C_m(\vect{r})$ is the rendered main camera ray.
% where $L_{ssim}$ is the structural similarity (SSIM) loss~\cite{DBLP:journals/tci/ZhaoGFK17}, $\hat C_b(\vect{r})$ is the rendered bokeh from the ultra-wide camera ray input, and $\hat C_m(\vect{r})$ is the rendered main camera ray.

{The final stage is to predict the blending mask. The goal is to fuse the high-quality details from the main camera and the sharp contents from the ultra-wide camera. The fusing process is integrated into the volume rendering process.}
The NeRF model trained by the main camera rays also outputs the blending mask $\eta$. For the training process, we still use the MSE loss
\begin{equation}
    \begin{aligned}
    \mathcal L_{fusion} =\sum_{r=1}^{N}\Vert C_{fuse}(\vect{r})-\hat{C}_f(\vect{r})\Vert_2^2\,,
    \end{aligned}
\end{equation}
where the ground truth $C_{fuse}$ is synthsized as  
\begin{equation}
    \begin{aligned}
    C_{fuse}=M_{blend}\cdot C_{w}+(1-M_{blend})\cdot C_{m}\,.
    \end{aligned}
\end{equation}
$M_{blend}$ is the defocus map $D_{defocus}$ with normalization to $0$ and $1$.

\noindent\textbf{Implementation Details.}
We implement our model by PyTorch~\cite{paszke2017automatic}. We use COLMAP~\cite{schonberger2016structure} to estimate the camera intrinsics and extrinsics, and these parameters are fixed during optimization. 
% During the training of ultra-wide and main rays, we sample xxx. 
{The smooth parameter for blur kernel weight distribution} $\beta$ is set as $4$ in our implementation of the kernel, {gamma correction parameter} $\gamma$ is set as $2$ . The focal distance $D_f$ is set to 0.5 for initialization, and the blur intensity $A$ is initialized with 5.
% For training of the fusion module, we sample a kernel as mentioned xxx, the size of the kernel is . 
We use the Adam optimizer~\cite{kingma2015adam} with a learning rate of $0.0005$, and decays exponentially to one-tenth for every $250000$ steps. Training a model takes around one day per scene using one NVIDIA GeForce GTX 3090 GPU for each $1008\times756$ frame. 

\begin{figure*}[htp]
  \centering
  \setlength{\abovecaptionskip}{3pt}
    \setlength{\belowcaptionskip}{0pt}
  \includegraphics[width=\linewidth]{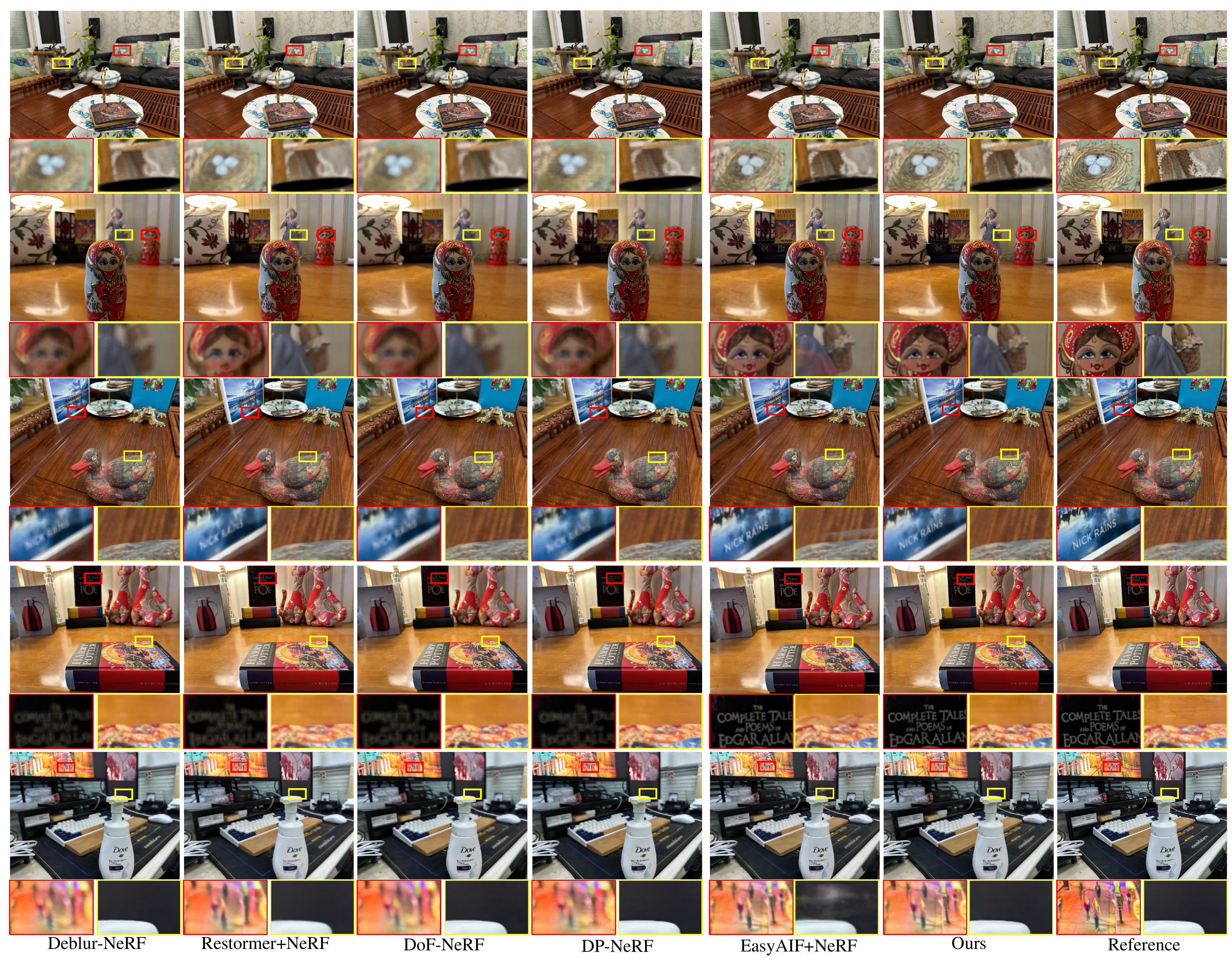}
  \caption{{Comparison with baselines on the smartphone dataset.} {The input pair of our method is the \textbf{foreground-focused} main image and the ultra-wide image. The inputs of baselines are only the main-camera images.} The red frame shows that our \ourmethod restores severe blur in the background that other methods fail to resolve. The yellow frame shows that EasyAIF~\cite{luo2022point} produces artifacts on the edges of the foreground object.}
  % The input pair is the main image and the ultra-wide image. 
  % Please zoom in to see the details.}
  \label{fig:baseline_fg}
\end{figure*}

\begin{figure*}[htp]
  \centering
  \setlength{\abovecaptionskip}{3pt}
    \setlength{\belowcaptionskip}{0pt}
  \includegraphics[width=\linewidth]{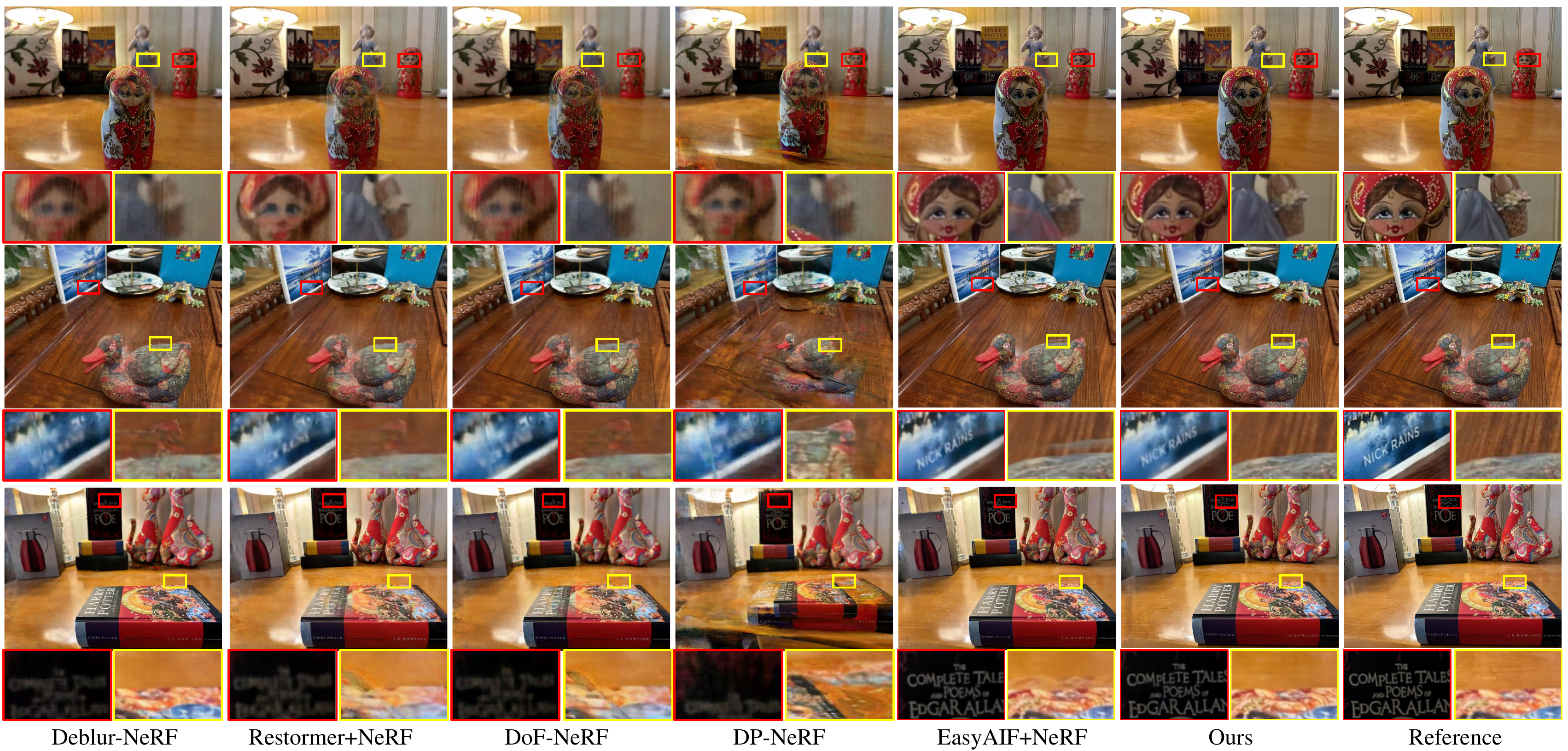}
  \caption{{Comparison with baselines on the smartphone dataset. All the methods are trained on the \textbf{foreground-focused} main image and the ultra-wide image. The red frame shows that our \ourmethod restores severe blur in the background that other methods fail to resolve. The yellow frame shows that EasyAIF~\cite{luo2022point} produces artifacts on the edges of the foreground object.}}
  \label{fig:baseline_fg_mainwide}
\end{figure*}
\begin{figure*}[htp]
  \centering
  \setlength{\abovecaptionskip}{3pt}
    \setlength{\belowcaptionskip}{0pt}
  \includegraphics[width=\linewidth]{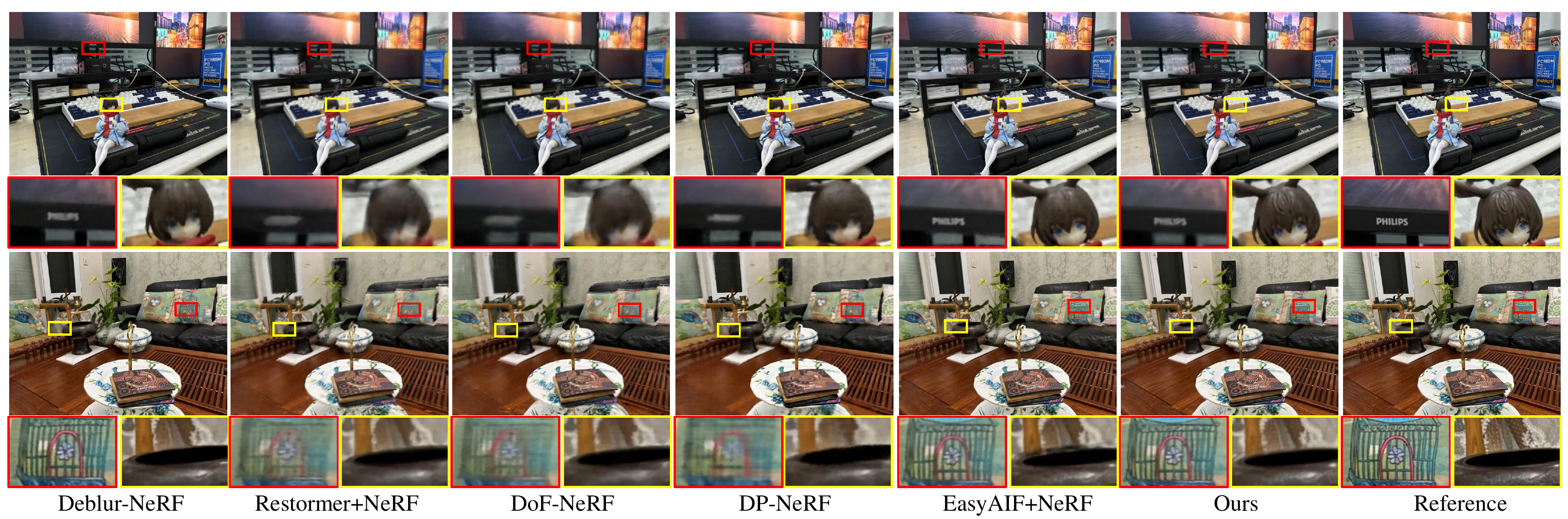}
  \caption{{Comparison with baselines on the smartphone dataset. All the methods are trained on the \textbf{foreground-focused} main image and the ultra-wide image aligned with the unified intrinsic. The red frame shows that our \ourmethod restores severe blur in the background that other methods fail to resolve. Deblur-NeRF produces more details with the aligned inputs, but it produces more misalignment than \ourmethod, for example, the hair/keyboard position of the doll scene and the birdcage position of the plate scene.}}
  \label{fig:baseline_fg_mainwide_match}
\end{figure*}

\begin{figure*}
  \centering
  \setlength{\abovecaptionskip}{3pt}
    \setlength{\belowcaptionskip}{0pt}
  \includegraphics[width=\linewidth]{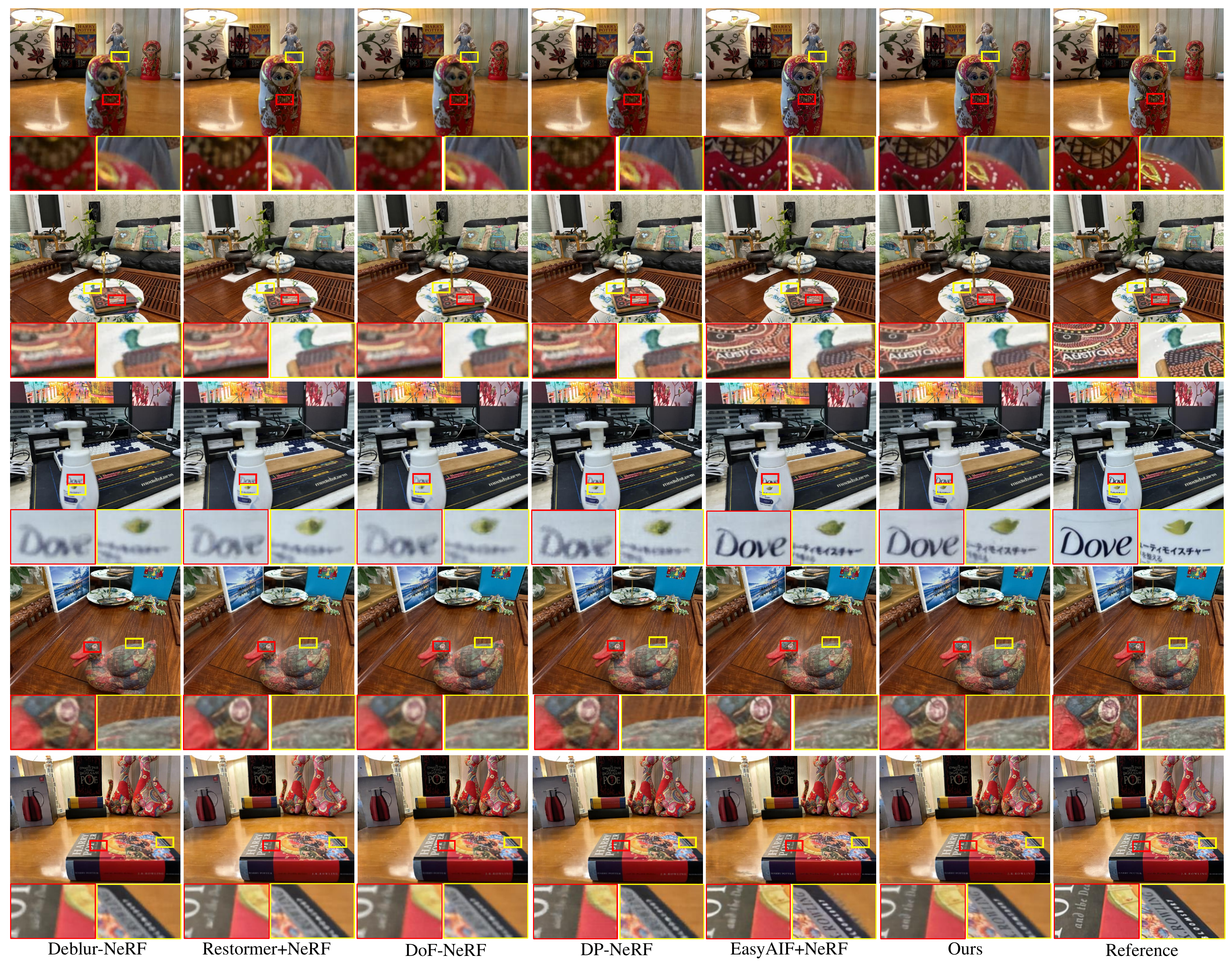}
  \caption{{Comparison with baselines on the smartphone dataset.} {The input pair of \ourmethod is the \textbf{background-focused} main image and the ultra-wide image. The inputs of baselines are only the main-camera images.} Compared with the baselines, our proposed \ourmethod restores sharper details and produces fewer artifacts on the edges.}
  \label{fig:baseline_bg}
\end{figure*}
%%%%%%%%%%%%%%%%%%%%%%%%%%%%%%%%%%%%%%%%%%%%%%%%%%%%%%%%%%%%%%%%

%%%%%%%%%%%%%%%%%%%%%%%%%%%%%%%%%%%%%%%%%%%%%%%%%%%%%%%%%%%%%%%%

\section{Experiments}
All the experiments are conducted on our captured smartphone dataset. The smartphone dataset consists of 7 scenes: \textit{doll}, \textit{bear}, \textit{plate}, \textit{book}, \textit{dove}, \textit{duck}, and \textit{russiandoll}. We capture 24 views per scene, and we split these views into 21 views for training and 3 for evaluation. Each view contains an image triplet, including a wide DoF ultra-wide image and two shallow DoF main camera images focused on the foreground and background respectively. We compare the 7 scenes with the main camera focused on 2 focal planes, which sums up to 14 sets of results. We set the resolution to $1008\times 756$ for training and evaluation.
From the experiments on the smartphone dataset, {we show that \ourmethod can deal with multi-view main-camera consistent defocus blur assisted by ultra-wide lens to achieve state-of-the-art performance}, and our approach can be extended into DoF applications such as refocusing and split diopter. 
% From the experiments on the smartphone dataset, we show that \ourmethod can deal with multi-view consistent defocus blur to achieve state-of-the-art performance, and our approach can be extended into DoF applications such as refocusing and split diopter. 

\begin{table*}[htp]
	\centering
	\caption{Quantitative results on our dataset. The input pair is the \textbf{foreground-focused} main image and the ultra-wide image. The ultra-wide alignment is on the image level. 
	The best performance is in \textbf{boldface}, and the second best is \underline{underlined}. EasyAIF requires an extra dataset for pre-training.} 
	\resizebox{1.0\linewidth}{!}
	{
    \setlength{\tabcolsep}{3pt}
	\renewcommand\arraystretch{1.0}
	\begin{NiceTabular}{l|ccc|ccc|ccc|ccc}
		\toprule
		\multicolumn{1}{l}{\multirow{2}{*}[-0.5ex]{Method}} &
  \multicolumn{3}{c}{Doll} & \multicolumn{3}{c}{Bear}& \multicolumn{3}{c}{Book}& \multicolumn{3}{c}{Dove} \\
		\cmidrule{2-13}
		~ &  PSNR$\uparrow$ &  SSIM$\uparrow$&LPIPS$\downarrow$ & PSNR$\uparrow$ & SSIM$\uparrow$ & LPIPS$\downarrow$ &PSNR$\uparrow$ &  SSIM$\uparrow$ & LPIPS$\downarrow$ &PSNR$\uparrow$ &  SSIM$\uparrow$&LPIPS$\downarrow$ \\
        \midrule
		\midrule
		Deblur-NeRF~\cite{ma2022deblur} & 22.56 & 0.741&0.277& 24.60 & 0.748&0.371 & 24.09& 0.727& 0.309&21.63&0.718&0.302\\
		{~\cite{ma2022deblur}+Ultra-wide} & 20.63 & 0.622&0.402& 22.68 & 0.671&0.485 & 21.59& 0.645& 0.388&19.34&0.594&0.432\\
		{~\cite{ma2022deblur}+Ultra-wide (Aligned)} & 20.78 & 0.632&0.378& 22.76 & 0.693&0.350 & 20.76& 0.659& 0.270&19.34&0.602&0.450\\
		DoF-NeRF~\cite{dofnerf}& 22.95 & 0.747 &  0.299& 24.71 &0.749 & 0.409& 24.48&0.733&0.324&21.66&0.709&0.341\\
		{~\cite{dofnerf}+Ultra-wide}& 21.17 & 0.635 &  0.420& 22.95&0.676 & 0.503& 22.43&0.645&0.440&19.49&0.588&0.466\\
		{~\cite{dofnerf}+Ultra-wide (Aligned)}& 21.13 & 0.643 &  0.427& 22.77&0.692 & 0.508& 19.81&0.621&0.438&18.61&0.540&0.489\\
        DP-NeRF~\cite{Lee_2023_CVPR}& 22.73 & 0.735 &0.287& 24.74 &0.744 & 0.394& 24.28 &0.727&0.310&21.72&0.710&0.312\\
        {~\cite{Lee_2023_CVPR}+Ultra-wide}&14.03 & 0.409 &0.517& 17.51 &0.547 & 0.586& 13.14&0.391&0.617&18.80&0.564&0.451\\
        {~\cite{Lee_2023_CVPR}+Ultra-wide (Aligned)}&17.48 & 0.549 &0.355& 22.59 &0.688 & 0.431& 18.90&0.652&0.364&20.60&0.644&0.360\\
        Restormer~\cite{zamir2022restormer}+ NeRF & 22.85 & 0.759&0.282& 25.03& 0.766& 0.350 &24.44 & 0.750 & 0.295&21.98&0.734&0.316\\
        {~\cite{zamir2022restormer}+ NeRF+Ultra-wide} &21.17 & 0.649&0.410& 23.06& 0.689& 0.476 &22.35 & 0.657 & 0.421&19.61&0.603&0.463\\
        {~\cite{zamir2022restormer}+ NeRF+Ultra-wide (Aligned)} &19.74 & 0.616&0.431& 22.38& 0.688& 0.504 &19.51 & 0.612 & 0.458&20.60&0.615&0.457\\
        EasyAIF~\cite{luo2022point}+NeRF&\underline{23.85} & \underline{0.784} &  \underline{0.240}& \textbf{25.33}&\underline{0.784} & \underline{0.330}& \underline{25.25} &\underline{0.780}&\underline{0.251}&\textbf{22.84}&\underline{0.763}&\underline{0.255}\\
		\ourmethod (Ours)& \textbf{24.22} & \textbf{0.787}&\textbf{0.228} & \underline{25.12}& \textbf{0.788} &\textbf{0.291}& \textbf{25.32}& \textbf{0.788} & \textbf{0.220}& \underline{22.72}&\textbf{0.765}&\textbf{0.222}\\
        \midrule
        \multicolumn{1}{l}{\multirow{2}{*}[-0.5ex]{Method}} &
  \multicolumn{3}{c}{Duck} & \multicolumn{3}{c}{Plate}& \multicolumn{3}{c}{Russiandoll}& \multicolumn{3}{c}{Average} \\
		\cmidrule{2-13}
  ~ &  PSNR$\uparrow$ &  SSIM$\uparrow$&LPIPS$\downarrow$ & PSNR$\uparrow$ & SSIM$\uparrow$ & LPIPS$\downarrow$ &PSNR$\uparrow$ &  SSIM$\uparrow$ & LPIPS$\downarrow$ &PSNR$\uparrow$ &  SSIM$\uparrow$&LPIPS$\downarrow$ \\
        \midrule
		\midrule
        Deblur-NeRF~\cite{ma2022deblur} & 22.79&0.654&\underline{0.284}&22.51&0.655&0.325&24.41&0.734&0.359& 23.23&0.711 &0.318 \\
        {~\cite{ma2022deblur}+Ultra-wide} & 21.61 & 0.522&0.497& 20.52 & 0.542&0.445 & 20.91& 0.618& 0.444&21.04&0.602&0.442\\
        {~\cite{ma2022deblur}+Ultra-wide (Aligned)} & 21.50 & 0.530&0.494& 20.60 & 0.560&0.430 & 20.95& 0.628& 0.440&20.96&0.615&0.402\\
		DoF-NeRF~\cite{dofnerf}& 22.80&0.656&0.318&23.66&0.663&0.366&24.52&0.737&0.386&23.54 & 0.713 & 0.349\\
          {~\cite{dofnerf}+Ultra-wide}& 21.56 & 0.522 &  0.514& 21.14 &0.552 & 0.475& 22.89&0.666&0.470&21.66&0.612&0.470\\
          {~\cite{dofnerf}+Ultra-wide (Aligned)}& 21.43 & 0.511 &  0.598& 20.97 &0.547 & 0.488& 22.76&0.690&0.486&21.07&0.606&0.491\\
        DP-NeRF~\cite{Lee_2023_CVPR}&22.48 & 0.638 & 0.293& 21.63 &0.604 & 0.339&24.18 &0.727&0.350&23.11&0.698&0.326\\
         {~\cite{Lee_2023_CVPR}+Ultra-wide}& 16.21 & 0.334 &0.619& 14.51&0.361 & 0.532& 16.73 &0.500&0.553&15.85&0.444&0.554\\
         {~\cite{Lee_2023_CVPR}+Ultra-wide (Aligned)}& 19.43 & 0.498 &0.514& 18.76&0.542 & 0.437& 21.02 &0.605&0.489&19.83&0.597&0.421\\
		Restormer~\cite{zamir2022restormer}+ NeRF &23.09&\underline{0.664}&0.309&23.16&0.690&0.321&24.68&0.754&0.321& {23.60} & 0.731 &0.313\\
      {~\cite{zamir2022restormer}+ NeRF+Ultra-wide} & 21.68 & 0.524&0.518& 21.21& 0.571& 0.436 &22.96& 0.684 & 0.426&21.72&0.625&0.450\\
      {~\cite{zamir2022restormer}+ NeRF+Ultra-wide (Aligned)} & 20.56 & 0.522&0.542& 19.20& 0.509& 0.500 &21.54& 0.611 & 0.502&20.50&0.596&0.485\\
        EasyAIF~\cite{luo2022point}+NeRF&\underline{23.50} & 0.638 & 0.293& \underline{23.77} &\underline{0.719} & \underline{0.278}& \underline{25.15}&\underline{0.787}&\underline{0.269}&\underline{24.24}&\underline{0.751}&\underline{0.274} \\
		\ourmethod (Ours)& \textbf{23.54}&\textbf{0.690}&\textbf{0.268}&\textbf{24.05}&\textbf{0.731}&\textbf{0.239}&\textbf{25.19}&\textbf{0.796}&\textbf{0.241}& \textbf{24.31}&\textbf{0.764}&\textbf{0.244}\\
    \bottomrule
	\end{NiceTabular}
	}
	\label{tab:baseline_fg}
\end{table*}

\begin{table*}[htp]
	\centering
	\caption{Quantitative results on our dataset. The input pair is the \textbf{background-focused }main image and the ultra-wide image.
	The best performance is in \textbf{boldface}, and the second best is \underline{underlined}. 
 % EasyAIF requires an extra dataset for pre-training. 
    The foreground object takes up much less space than the background, so the background-focused main image has less blur and is easier to fix, and the metric differences between these methods are less obvious in numbers. Under this circumstance, \ourmethod is still on par with EasyAIF which requires an extra dataset for pre-training.
    Although Deblur-NeRF achieves better LPIPS, we urge readers to view the qualitative results and the supplementary video where \ourmethod achieves better multi-view consistency and restores sharper details than Deblur-NeRF.} 
	\resizebox{1.0\linewidth}{!}
	{
    \setlength{\tabcolsep}{3pt}
	\renewcommand\arraystretch{1.0}
	\begin{NiceTabular}{l|ccc|ccc|ccc|ccc}
		\toprule
		\multicolumn{1}{l}{\multirow{2}{*}[-0.5ex]{Method}} &
  \multicolumn{3}{c}{Doll} & \multicolumn{3}{c}{Bear}& \multicolumn{3}{c}{Book}& \multicolumn{3}{c}{Dove} \\
		\cmidrule{2-13}
		~ &  PSNR$\uparrow$ &  SSIM$\uparrow$&LPIPS$\downarrow$ & PSNR$\uparrow$ & SSIM$\uparrow$ & LPIPS$\downarrow$ &PSNR$\uparrow$ &  SSIM$\uparrow$ & LPIPS$\downarrow$ &PSNR$\uparrow$ &  SSIM$\uparrow$&LPIPS$\downarrow$ \\
        \midrule
		\midrule
        % naive NeRF~\cite{mildenhall2020nerf}& 25.24 & 0.717 &0.422 &   25.72 & 0.722 &0.419& 25.48& 0.719 & 0.421\\
		Deblur-NeRF~\cite{ma2022deblur} &24.18 & 0.806&\textbf{0.140}& 23.98 & 0.743&\textbf{0.218} & 25.15& {0.802}& \underline{0.188}&23.26&\underline{0.806}&\textbf{0.141}\\
  DoF-NeRF~\cite{dofnerf}& 22.92 & 0.744 &  0.300& {24.92}&{0.758} & {0.319}& {25.39}&0.798&0.224&{23.34}&0.800&0.189\\
  DP-NeRF~\cite{Lee_2023_CVPR}&24.29 & 0.792 & \underline{0.163}& 23.89 &0.739 & \underline{0.264}&25.52 &0.794&0.206&22.90&0.776&\underline{0.175}\\
		Restormer~\cite{zamir2022restormer}+ NeRF & 24.05 & 0.791&0.219& 24.65& 0.751& 0.365 &25.09 & 0.785 & 0.258 &23.08&0.780&0.232\\
  EasyAIF~\cite{luo2022point}+NeRF&\textbf{25.97} & \underline{0.827} & 0.167& \textbf{25.37} &\textbf{0.777} & {0.305}& \underline{25.84}&\underline{0.814}&{0.195}&\underline{23.56}&{0.795}&{0.202} \\
		\ourmethod (Ours)& \underline{25.80} & \textbf{0.828}&{0.169}& \underline{25.10}& \underline{0.760} &0.311& \textbf{26.51}& \textbf{0.824} & \textbf{0.182}& \textbf{23.82}&\textbf{0.808}&\underline{0.175}\\
        \midrule
        \multicolumn{1}{l}{\multirow{2}{*}[-0.5ex]{Method}} &
  \multicolumn{3}{c}{Duck} & \multicolumn{3}{c}{Plate}& \multicolumn{3}{c}{Russiandoll}& \multicolumn{3}{c}{Average} \\
		\cmidrule{2-13}
  ~ &  PSNR$\uparrow$ &  SSIM$\uparrow$&LPIPS$\downarrow$ & PSNR$\uparrow$ & SSIM$\uparrow$ & LPIPS$\downarrow$ &PSNR$\uparrow$ &  SSIM$\uparrow$ & LPIPS$\downarrow$ &PSNR$\uparrow$ &  SSIM$\uparrow$&LPIPS$\downarrow$ \\
        \midrule
		\midrule
        Deblur-NeRF~\cite{ma2022deblur} & 23.84&0.680&\underline{0.225}&23.14&0.743&\textbf{0.167}&{24.52}&{0.784}&{0.212}& 24.01&{0.766} &\textbf{0.184 }\\
        DoF-NeRF~\cite{dofnerf}&\underline{24.19}&{0.692}&0.242&{23.66}&\underline{0.752}&{0.203}&24.06&0.770&0.265&24.07 & 0.759 & 0.249\\
        DP-NeRF~\cite{Lee_2023_CVPR}&23.36 & 0.644 & 0.252&22.99 &0.718 & 0.206&23.80&0.768&\textbf{0.203}&23.82&0.747&{0.210}\\
		Restormer~\cite{zamir2022restormer}+ NeRF &23.91&0.691&0.266&23.01&0.729&0.274&23.63&0.766&0.293& 23.92 & 0.756&0.272\\
  EasyAIF~\cite{luo2022point}+NeRF&\underline{24.19} & \underline{0.701} & 0.247& \underline{23.77} &{0.740} & {0.216}& \textbf{25.22}&\textbf{0.815}&{0.215}&\underline{24.85}&\underline{0.781}&{0.221} \\
		\ourmethod (Ours)& \textbf{24.54}&\textbf{0.717}&\textbf{0.220}&\textbf{24.25}&\textbf{0.764}&\underline{0.194}&\underline{24.72}&\underline{0.797}&\underline{0.209}& \textbf{24.96}&\textbf{0.785}&\underline{0.209}\\
    \bottomrule
	\end{NiceTabular}
	}
	\label{tab:baseline_bg}
\end{table*}
\subsection{Baseline Approaches}
We choose state-of-the-art baselines that are closely related to AiF NeRF synthesis, including two types of approaches: 1) NeRF for defocused inputs~\cite{ma2022deblur,dofnerf,Lee_2023_CVPR} and 2) an image-space baseline that uses single-view image deblurring~\cite{zamir2022restormer,luo2022point} then trains the NeRF with the deblurred result. It is noted that DC2~\cite{alzayer2023defocuscontrol} also provides a baseline for deblurring. Since DC2 does not provide source code and dataset for training, we use the concurrent work EasyAIF~\cite{luo2022point} that is very similar to DC2 for comparison.

\noindent\textbf{Deblur-NeRF~\cite{ma2022deblur}} is a NeRF method with a deformable sparse kernel to handle blurry inputs.

\noindent\textbf{DoF-NeRF~\cite{dofnerf}} is a NeRF method with controllable DoF rendering. 

\noindent\textbf{DP-NeRF~\cite{Lee_2023_CVPR}} is a NeRF method that utilizes physical scene priors to deblur the multi-view inputs. 

\noindent\textbf{Restormer~\cite{zamir2022restormer}} achieves image deblurring using Transformer. The deblurring model is pre-trained on a large dataset~\cite{abuolaim2020defocus}. 

\noindent\textbf{EasyAIF~\cite{luo2022point}} achieves single-view image deblurring by introducing the ultra-wide camera to assist the main camera, similar to DC2~\cite{alzayer2023defocuscontrol}. The deblurring model is pre-trained on a multi-camera dataset.

\begin{table*}[htp]
	\centering
	\caption{{Quantitative results on synthetic datasets. The best performance is in \textbf{boldface}, and the second best is \underline{underlined}.}} 
	\resizebox{1.0\linewidth}{!}
	{
    \setlength{\tabcolsep}{3pt}
	\renewcommand\arraystretch{1.0}
	\begin{NiceTabular}{l|ccc|ccc|ccc|ccc}
		\toprule
		\multicolumn{1}{l}{\multirow{2}{*}[-0.5ex]{Method}} &
  \multicolumn{3}{c}{Stadium} & \multicolumn{3}{c}{Tenhag}& \multicolumn{3}{c}{LAC}& \multicolumn{3}{c}{Average} \\
		\cmidrule{2-13}
		~ &  PSNR$\uparrow$ &  SSIM$\uparrow$&LPIPS$\downarrow$ & PSNR$\uparrow$ & SSIM$\uparrow$ & LPIPS$\downarrow$ &PSNR$\uparrow$ &  SSIM$\uparrow$ & LPIPS$\downarrow$ &PSNR$\uparrow$ &  SSIM$\uparrow$&LPIPS$\downarrow$ \\
        \midrule
		\midrule
        % naive NeRF~\cite{mildenhall2020nerf}& 25.24 & 0.717 &0.422 &   25.72 & 0.722 &0.419& 25.48& 0.719 & 0.421\\
		Deblur-NeRF~\cite{ma2022deblur} &\textbf{25.38} & 0.843&\underline{0.190}& 24.00 & 0.702&0.275 & 21.13& 0.684& 0.395&23.50&0.743&{0.287}\\
  DoF-NeRF~\cite{dofnerf}& 24.56 & \underline{0.848} &  0.201& 24.16&0.718 & 0.266& 21.11&0.684&0.416&23.28&0.750&0.294\\
		Restormer~\cite{zamir2022restormer}+ NeRF & 23.73 & 0.847&0.198& 24.22& 0.716& \underline{0.261} &\underline{21.59} & {0.714} & \underline{0.361} &23.18&0.759&\underline{0.273}\\
  EasyAIF~\cite{luo2022point}+NeRF&\underline{24.63}& 0.845 &0.210 & \underline{24.48} &\underline{0.720}& 0.266& {21.45}&\underline{0.720}&{0.393}&\underline{23.52}&\underline{0.762}&{0.290} \\
		\ourmethod (Ours)& 24.53 & \textbf{0.858}&\textbf{0.178}& \textbf{24.53}& \textbf{0.737} &\textbf{0.254}& \textbf{22.65}& \textbf{0.748} & \textbf{0.315}& \textbf{23.90}&\textbf{0.781}&\textbf{0.249}\\
  \bottomrule
	\end{NiceTabular}
	}
	\label{tab:synthetic}
\end{table*}

\begin{figure*}
    \setlength{\abovecaptionskip}{3pt}
    \setlength{\belowcaptionskip}{5pt}
  \includegraphics[width=\textwidth]{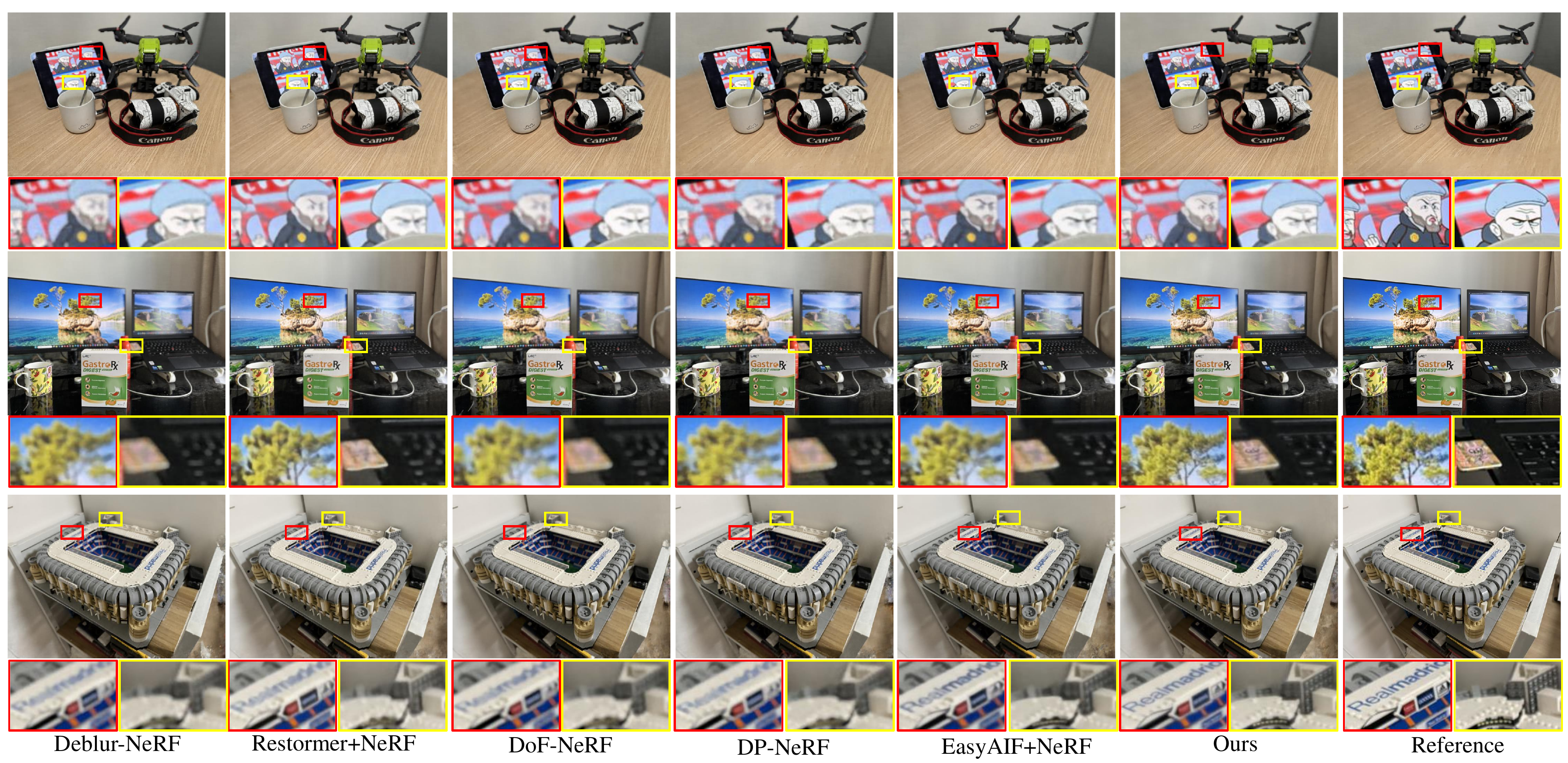}
  \caption{{Comparison with baselines on the synthetic dataset. Compared with the baselines, our proposed DC-NeRF restores sharper details.}
  }
  \label{fig:synthetic}
\end{figure*}
% wzj0503
% \begin{figure}[htp]
%   \centering
%   \setlength{\abovecaptionskip}{3pt}
%     \setlength{\belowcaptionskip}{0pt}
%   \includegraphics[width=\linewidth]{fig/ablation_component.pdf}
%   \caption{Ablation study on our use of the confidence map and the defocus-aware fusion module.}
%   \label{fig:ablation_component}
% \end{figure}

% wzj0503
% \begin{figure*}[htp]
%   \centering
%   \setlength{\abovecaptionskip}{3pt}
%     \setlength{\belowcaptionskip}{0pt}
%   \includegraphics[width=\linewidth]{fig/baseline.pdf}
%   \caption{Comparison with baselines on the smartphone dataset. Note that our method restores severe blur in the background that other methods fail to resolve.}
%   % The input pair is the main image and the ultra-wide image. 
%   % Please zoom in to see the details.}
%   \label{fig:baseline}
% \end{figure*}
\subsection{Qualitative Results}\label{quality}
% We show intermediate results in Fig.~\ref{fig:intermediate} to show how we synthesize an AiF novel view $I_{AiF}$ from the dual-camera input views. We apply image alignment to match $I_w$ to $I_m$, and output the aligned $I_{w,a}$. The defocus-aware fusion module predicts the defocus parameters, and they are used with the predicted disparity map to calculate a defocus map to fuse the two cameras. The 3D blending weight field is supervised by the fused result and it generates AiF novel views and their corresponding fusion masks.
% \begin{figure}
%   \centering
%   \setlength{\abovecaptionskip}{3pt}
%     % \setlength{\belowcaptionskip}{0pt}
%   \includegraphics[width=0.95\linewidth]{fig/intermediate.pdf}
%   \caption{Intermediate results of \ourmethod.}
%   \label{fig:intermediate}
% \end{figure}
We show qualitative comparisons in Fig.~\ref{fig:baseline_fg} and Fig.~\ref{fig:baseline_bg}. One can observe: 1) Deblur-NeRF~\cite{ma2022deblur}, DP-NeRF~\cite{Lee_2023_CVPR}, and DoF-NeRF~\cite{dofnerf} can not deal with the consistent defocus blur; 2) the image-space single-camera deblurring method~\cite{zamir2022restormer} fails to recover sharp details from large blur;
3) Our method \ourmethod outperforms these baselines in generating AiF novel views;
4) \ourmethod outperforms EasyAIF~\cite{luo2022point} by a margin because EasyAIF focuses on single-view deblurring and restores some sharp details for each view, but it does not take multi-view consistency into account. Therefore, EasyAIF produces more artifacts on the edges. Furthermore, noted that EasyAIF and a similar method DC2~\cite{alzayer2023defocuscontrol} both require an extra dataset for training, which introduces additional training costs.
{Additionally, we train the baseline methods on the main/ultra-wide image pair. Specifically, we conduct two separate experiments on (1) the unaligned ultra-wide and (2) the ultra-wide image aligned on image level. However, as shown in Fig.~\ref{fig:baseline_fg_mainwide} and Fig.~\ref{fig:baseline_fg_mainwide_match}, the baselines generate worse visual results.} 
{We list the possible reasons: (1) {the NeRF model is not capable of learning a feasible representation from multi-intrinsic inputs, and the model only converges well when the ray casting is fixed in the rendering process. Although we also align the ultra-wide camera on image level, misalignment still exists due to COLMAP deviations and unknown distortion;} (2) the predicted poses of different cameras from COLMAP can be fragile with sparse views. Multi-camera systems require precise calibration to ensure that the relative positions and orientations of the cameras are accurate. Any calibration errors can lead to severe inaccuracies in the reconstructed scene; (3) directly providing sharp contents from extra views does not help to deblur the consistent blurred main-camera views. The consistent blur means no supervision for the rays from the main camera view, and it is difficult to transfer the sharp content from ultra-wide ray to correspond with the blurred one in the main ray. Since it is difficult to model the defocus blur from a multi-camera setting, we apply alignment in our pipeline and switch the task to learning the fusion of sharp high-quality details from blur model.}

{We also collect three synthetic scenes and conduct experiments on these data. As shown in Fig.~\ref{fig:synthetic}, \ourmethod still performs better than the baselines.}

\ourmethod is able to restore the sharp details by transferring the deep DoF from the ultra-wide camera, and the high-quality region of the main camera is preserved. 
% To demonstrate our method's superiority, we also conduct experiments on input data with inconsistent defocus blur, where the focused object in each view is different. The results are shown in Fig.~\ref{fig:halfbaseline}.
Refer to the supplementary video for more results.

\subsection{Quantitative Results}
We compare \ourmethod with the five baseline methods quantitatively on our smartphone dataset.
% We split the evaluation dataset into two groups: the main images focused on foreground, and the main images focused on background. 
To evaluate the performance of these methods, we use PSNR, SSIM, and LPIPS~\cite{zhang2018unreasonable} as metrics. 

As shown in Table~\ref{tab:baseline_fg}, \ourmethod outperforms other state-of-the-art baselines. Specifically, \ourmethod outperforms EasyAIF~\cite{luo2022point} on most scenes and most test metrics. As mentioned in Sec.~\ref{quality}, \ourmethod utilizes multi-view consistency while EasyAIF is limited to single-view deblurring. EasyAIF focuses on deblurring individual views without considering consistency across viewpoints, resulting in artifacts on the edges. In contrast, DC-NeRF jointly optimizes all input views, which enforces coherence between views and allows higher quality novel views.
Additionally, we would like to emphasize that EasyAIF and the similar method DC2~\cite{alzayer2023defocuscontrol} require extra training data besides the images of the target scene. They pre-train networks on blurred/sharp image pairs before training NeRF on the specific scene. In comparison, \ourmethod is trained directly on the target data like all the other NeRF methods, avoiding the need for supplemental training data. This gives \ourmethod an advantage by reducing training cost.
{Similar to the qualitative experiments, we train the baseline methods on the main/ultra-wide image pair. Specifically, we conduct two separate experiments on (1) the unaligned ultra-wide and (2) the aligned ultra-wide image with the unified intrinsic. However, as shown in Tab.~\ref{tab:baseline_fg}, the baselines trained on the main/ultra-wide image pair generate worse visual results. We can also observe that although the rendered visualization on the aligned ultra-wide inputs is better than that of the unaligned one, but the misalignment between the rendered view and the ground truth hinders the improvement quantitatively.  }

In Table~\ref{tab:baseline_bg}, the metric differences between these methods are less apparent in numbers because the foreground object takes up much less space than the background, so the background-focused main image has less blur and is easier to fix. Under this circumstance, \ourmethod is still on par with EasyAIF which requires an extra dataset for pre-training. Although Deblur-NeRF achieves better LPIPS, we urge readers to view the qualitative results and the supplementary video where \ourmethod achieves better multi-view consistency and restores sharper details than Deblur-NeRF.
We also observe that the single-view image deblurring methods also perform slightly better than the NeRF models. We believe this is because both NeRF models tend to fail when the inputs have consistent defocus blur, and the single-view image deblurring method at least alleviates the defocus blur on each input view. 
The quantitative results of the synthetic dataset are shown in Tab~\ref{tab:synthetic}.
Refer to the supplementary video for more results.

% Restormer~\cite{zamir2021restormer} and DCSR~\cite{wang2021dual} are the second best approaches. We demonstrate that \ourmethod is superior in terms of image focused on the foreground as well as on the background. Reference-based super-resolution methods MASA-SR~\cite{lu2021masa} and DCSR perform well to super-resolve the already clear contents in the image, however, they fail to handle large defocus blur. Restormer and IFAN~\cite{lee2021iterative} are designed for defocus deblurring, but they do not fully  utilize the sharp contents in the reference image $I_w$.

% wzj0503
% \begin{figure}[htp]
\begin{figure}[t!]
  \centering
  \setlength{\abovecaptionskip}{3pt}
    \setlength{\belowcaptionskip}{0pt}
  \includegraphics[width=\linewidth]{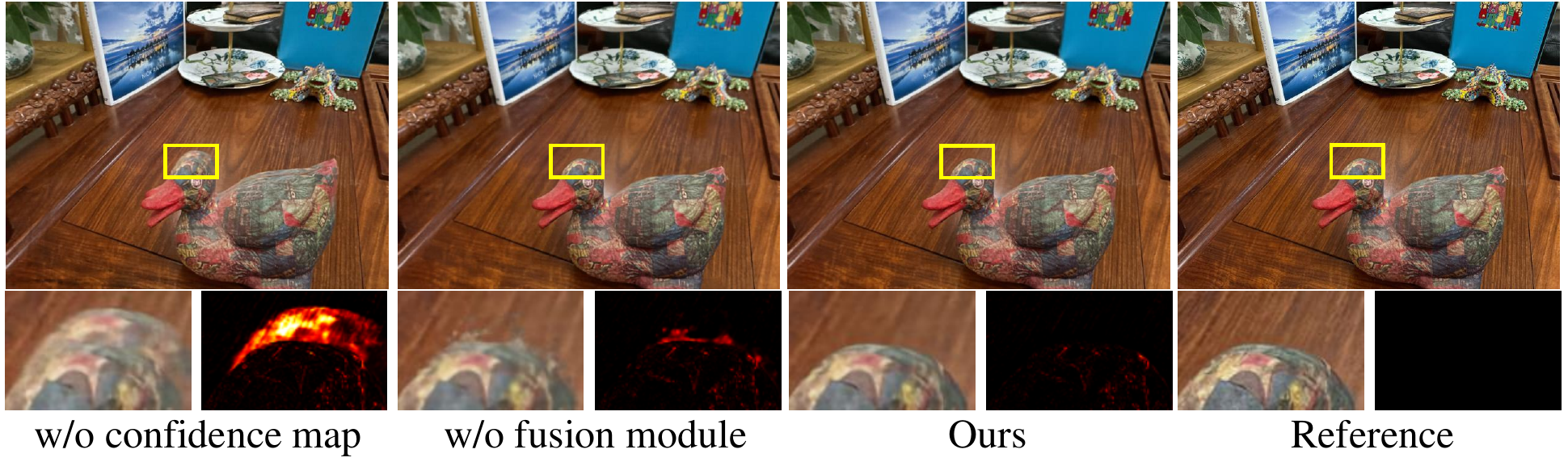}
  \caption{{Ablation study on framework components.} We compute the error map to show the difference, where darker regions indicate smaller errors.}
  \label{fig:ablation_component}
\end{figure}
\begin{figure}[t!]
  \centering
  \setlength{\abovecaptionskip}{3pt}
    \setlength{\belowcaptionskip}{0pt}
  \includegraphics[width=\linewidth]{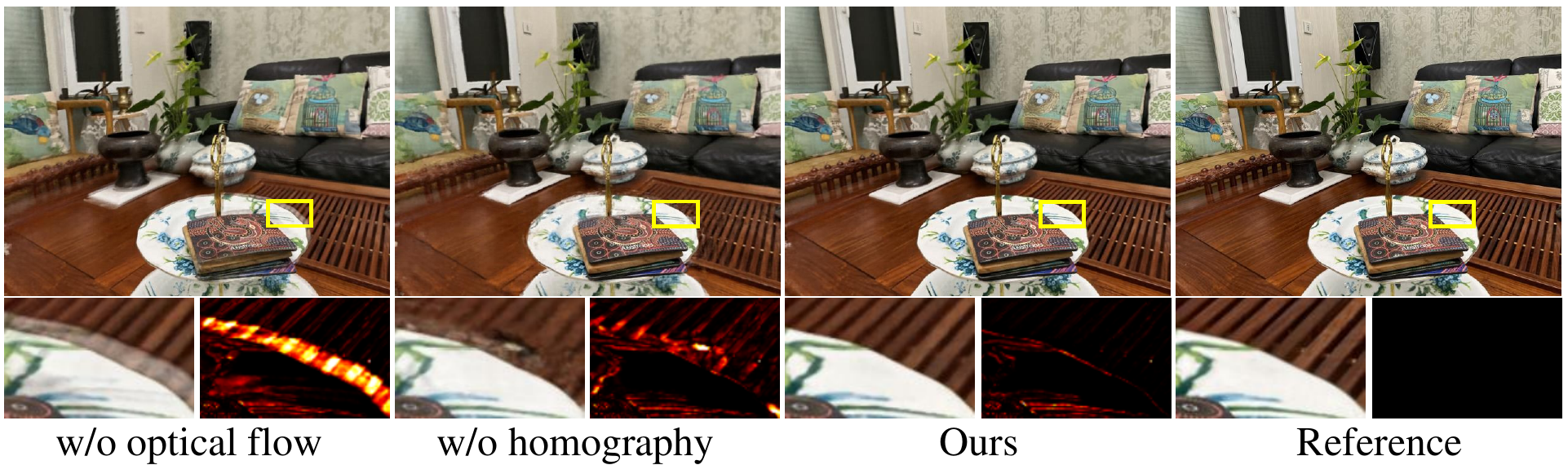}
  \caption{{Ablation study on our warping operations.} We compute the error map to show the difference, where darker regions indicate smaller errors.}
  \label{fig:ablation_alignment}
\end{figure}
% wzj0503
% \begin{figure}[htp]
\begin{figure}[t!]
  \centering
  \setlength{\abovecaptionskip}{3pt}
    \setlength{\belowcaptionskip}{0pt}
  \includegraphics[width=\linewidth]{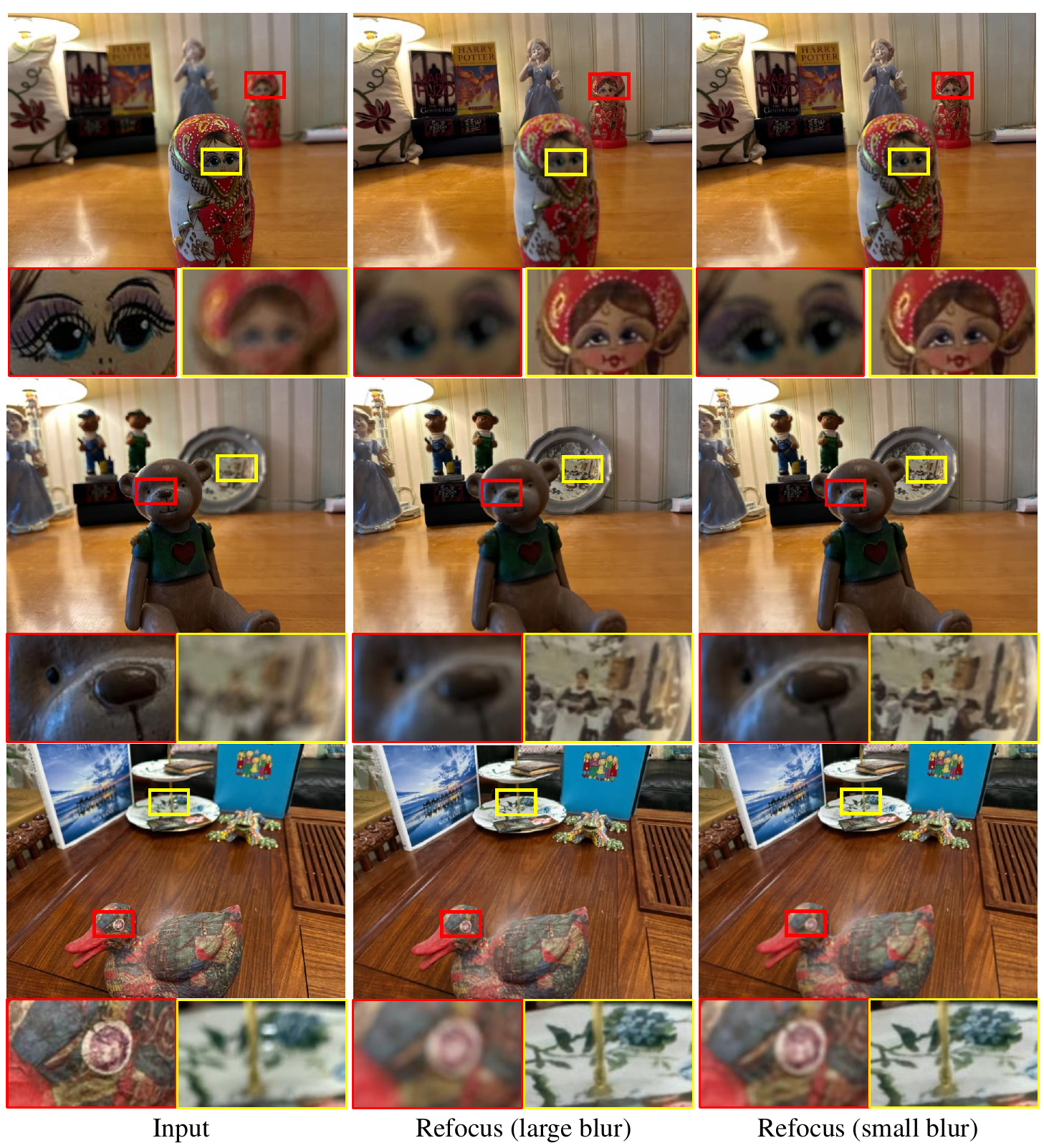}
  \caption{\textbf{Refocusing from \ourmethod.} The input view is focused on an object, the refocusing shifts the focal plane to another object of a different depth. We also show that the blur intensity can be adjusted.}
  \label{fig:refocus}
\end{figure}

\begin{figure}[t!]
  \centering
  \setlength{\abovecaptionskip}{3pt}
    \setlength{\belowcaptionskip}{0pt}
  \includegraphics[width=0.9\linewidth]{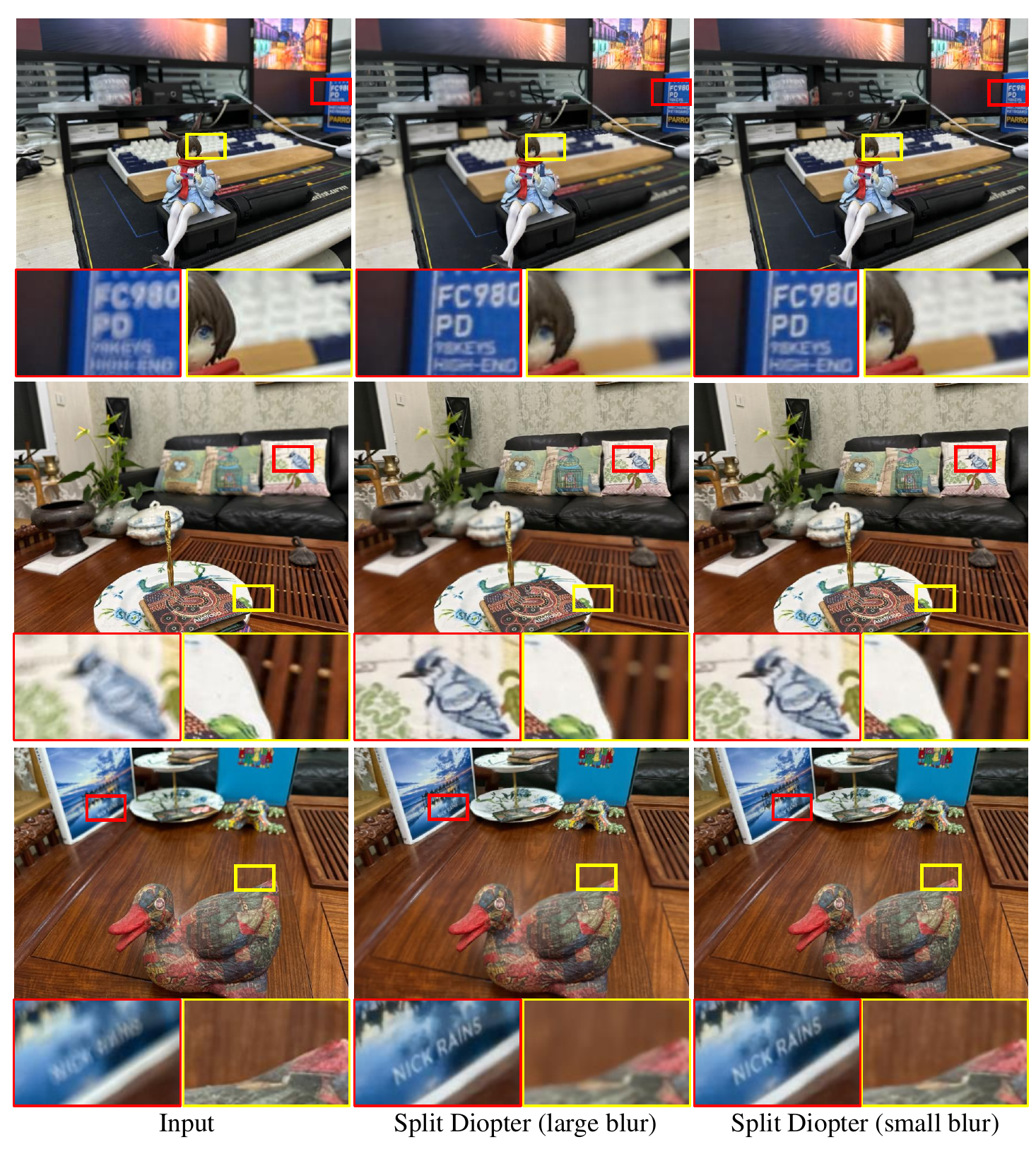}
  \caption{{Split diopter from \ourmethod.} We mimic the split diopter lens used in movies, where the foreground and background keep focused, and the region in between is blurred. We also show that the blur intensity can be adjusted.}
  \label{fig:split}
\end{figure}

\subsection{Ablation Study}
% We conduct ablation studies on the framework components. 
The main idea of \ourmethod is to exploit the ultra-wide camera for sharp content, and the challenge is to align and fuse the two inputs. Therefore, we first evaluate the effects of image alignment. We compare the performance in Table.~\ref{tab:spatial_ablation}. We show that image registration with homography and optical flow warping both improve our performance. Homography warping provides image-level alignment, and optical flow focuses on pixel-level warping.
% We show that the pixel-level optical flow alignment and the image-level homography warping are both necessary. 
% As shown in Tab~\ref{tab:spatial_ablation}, homography alignment and flow warping both improve our performance. Homography warping provides image-level alignment, and optical flow focuses on pixel-level warping.
% Note that if we only use optical flow for spatial alignment, despite that the pixel-level alignment is sufficient, the warped image still suffers from degraded quality due to the large gap between the main camera and the ultra-wide camera. 
We visualize the results in Fig.~\ref{fig:ablation_alignment}. 
% ablation放sm
% Refer to the supplementary material for qualitative results.
{We also evaluate the effects of the confidence map, the defocus-aware fusion module, and the loss terms $\mathcal L_{fusion}$, $\mathcal{L}_{\mathit{focus}}$. As shown in Table~\ref{tab:ablation} and Table~\ref{tab:ablation_loss}, our framework works best when both components and all the loss terms are used.}

The visualizations are shown in Fig.~\ref{fig:ablation_component}. 
The confidence map helps the ultra-wide radiance field to eliminate rays with artifacts for training. The defocus-aware fusion module provides a more accurate blending mask for the dual camera.
For the model without the defocus-aware fusion module, we use the disparity map as the blending mask. We show that \ourmethod performs better on the edges of objects from the visualizations. {For the model trained without $\mathcal L_{fusion}$ and the model trained without $\mathcal{L}_{\mathit{focus}}$, we still retain the defocus-aware fusion module, but the supervision on the rendered result is removed.}

\begin{figure}[!h]
  \centering
  \setlength{\abovecaptionskip}{3pt}
  \setlength{\belowcaptionskip}{-10pt}
  \includegraphics[width=0.9\linewidth]{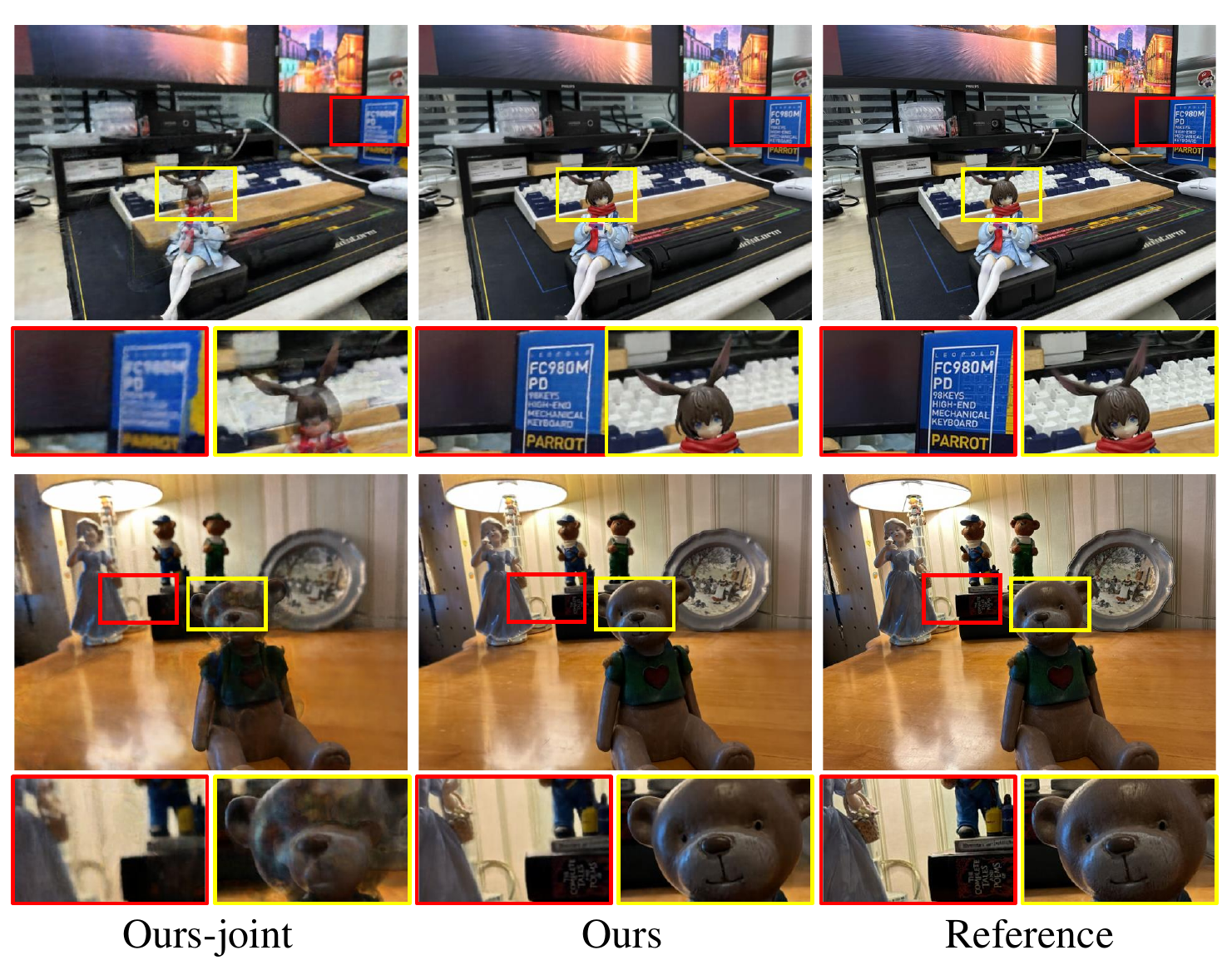}
  \caption{{We conduct experiments to compare our current pipeline with \ourmethod trained on a single NeRF with the defocus parameters predicted for the main cameras.}}
  \label{fig:joint}
\end{figure}

{To show the necessity of image alignment, we conduct experiments using the unaligned ultra-wide images as additional inputs, predicting defocus parameters for main cameras. The result is shown in Tab.~\ref{tab:trainwide_dof} and Fig.~\ref{fig:joint}, we prove that our aligned-ultra-wide method performs better than directly using unaligned ultra-wide inputs qualitatively and quantitatively. We think our pipeline works better because the defocus parameters predictions are difficult if the defocus blur is consistent for main-camera views. Although the additional ultra-wide ray is available, the multi-camera setting hinders the correspondence between the sharp ultra-wide ray and the blurred main ray. Therefore, we can observe in the figure that the blurred regions are not restored. Furthermore, NeRF models are not designed for multi-camera poses, and the varying intrinsic parameters result in poor training convergence.}

\begin{table}
\centering
  \caption{Ablation study on warping operations in spatial alignment. The best performance is in \textbf{boldface}. }
  \begin{tabular}{lcccc}
    \toprule
    Scene& Warping methods&PSNR$\uparrow$ &  SSIM$\uparrow$ & LPIPS$\downarrow$\\
    \midrule
    \multirow{3}*{Book}&Homography& 21.15& 0.662&0.352\\
    ~&Flow&22.54&0.690&0.304\\
    ~&Homography$+$Flow &\textbf{25.32}& \textbf{0.788}& \textbf{0.220}\\
    \midrule
    \multirow{3}*{Plate}&Homography& 19.45& 0.550&0.374\\
    ~&Flow&20.42&0.582&0.389\\
    ~&Homography$+$Flow &\textbf{24.05 }& \textbf{0.731}& \textbf{0.239}\\
    
  \bottomrule
\end{tabular}
\label{tab:spatial_ablation}
\end{table}

\begin{table}[!h]
	\centering
	\caption{Ablation study on framework components. The best performance is in \textbf{boldface}.} 
	\resizebox{1.0\linewidth}{!}
 {
    \setlength{\tabcolsep}{3pt}
	\renewcommand\arraystretch{1.0}
	% \begin{NiceTabular}{ccccccc}
	\begin{tabular}{ccccccc}
		\toprule
		% \multicolumn{1}{c}{Framework Components} &\multicolumn{3}{c}{Framework Components} & \multicolumn{3}{c}{Total} \\
		% \cmidrule{0-5}
		Scene & Confidence map&Fusion module& PSNR$\uparrow$ &  SSIM$\uparrow$ &LPIPS$\downarrow$  \\
        \midrule
        % wzj0503
		% \multirow{3}*{duck}& \checkmark & \times& 22.43& 0.67& 0.294\\
        \multirow{3}*{Duck}& $\checkmark$ & $\times$& 22.43& 0.670& 0.294\\
	  ~  & $\times$ &$\checkmark$ & 21.50& 0.657 &0.294\\
		~  &$\checkmark$ & $\checkmark$& \textbf{23.54}& \textbf{0.690}& \textbf{0.268}\\
        \midrule
		\multirow{3}*{Bear}& $\checkmark$& $\times$& 23.64& 0.771& 0.296\\
	  ~  &  $\times$ &$\checkmark$ & 22.55& 0.761 &0.328 \\
		~  &$\checkmark$ & $\checkmark$ & \textbf{25.12}& \textbf{0.788}& \textbf{0.291}\\
		\bottomrule
	% \end{NiceTabular}
	\end{tabular}
 }
	\label{tab:ablation}
\end{table}
\begin{table}[!h]
	\centering
	\caption{{Ablation study on the loss terms. The best performance is in \textbf{boldface}.}} 
	\resizebox{1.0\linewidth}{!}
 {
    \setlength{\tabcolsep}{3pt}
	\renewcommand\arraystretch{1.0}
	% \begin{NiceTabular}{ccccccc}
	\begin{tabular}{ccccccc}
		\toprule
		% \multicolumn{1}{c}{Framework Components} &\multicolumn{3}{c}{Framework Components} & \multicolumn{3}{c}{Total} \\
		% \cmidrule{0-5}
		Scene & $\mathcal L_{fusion}$&$\mathcal{L}_{\mathit{focus}}$  &PSNR$\uparrow$ &  SSIM$\uparrow$ &LPIPS$\downarrow$  \\
        \midrule
        \multirow{3}*{Duck}& $\checkmark$ & $\times$& 23.37& 0.677& 0.301\\
	  ~  & $\times$ &$\checkmark$ & 22.54& 0.679 &0.292\\
		~  &$\checkmark$ & $\checkmark$& \textbf{23.54}& \textbf{0.690}& \textbf{0.268}\\
        \midrule
		\multirow{3}*{Bear}& $\checkmark$& $\times$& 25.02&0.777 & 0.332\\
	  ~  &  $\times$ &$\checkmark$ & 24.69& 0.777 &0.330 \\
		~  &$\checkmark$ & $\checkmark$ & \textbf{25.12}& \textbf{0.788}& \textbf{0.291}\\
		\bottomrule
	% \end{NiceTabular}
	\end{tabular}
 }
	\label{tab:ablation_loss}
\end{table}
% \begin{table}[t]
% 	\centering
% 	\caption{{We conduct experiments of NeRF models trained on only ultra-wide images. We apply homography transform on the original rendered novel main-camera views of the baseline to compare the results. The original rendered results have large discrepancies with the main-camera view because the NeRF is not suitable for multi-camera. The best performance is in \textbf{boldface}.}} 
% 	\resizebox{1.0\linewidth}{!}
%  {
%     \setlength{\tabcolsep}{3pt}
% 	\renewcommand\arraystretch{1.0}
% 	% \begin{NiceTabular}{ccccccc}
% 	\begin{tabular}{cccccc}
% 		\toprule
% 		Scene &Method&PSNR$\uparrow$ &  SSIM$\uparrow$ &LPIPS$\downarrow$  \\
%         \midrule
%         \multirow{2}*{Doll}&NeRF+Ultra-wide&13.93& 0.446& 0.459\\
% 	  ~  & Ours& \textbf{24.22}& \textbf{0.787} &\textbf{0.228}\\
%         \midrule
% 		\multirow{2}*{Bear}&NeRF+Ultra-wide& 15.55&0.570 & 0.534\\
% 	  ~  &Ours&  \textbf{25.12}& \textbf{0.788} &\textbf{0.291} \\
% 		\bottomrule
% 	% \end{NiceTabular}
% 	\end{tabular}
%  }
% 	\label{tab:trainwide}
% \end{table}
\begin{table}[!h]
	\centering
	\caption{{We conduct experiments of \ourmethod trained on a single NeRF and defocus parameters predicted for the main cameras. The best performance is in \textbf{boldface}.}} 
	\resizebox{1.0\linewidth}{!}
 {
    \setlength{\tabcolsep}{3pt}
	\renewcommand\arraystretch{1.0}
	% \begin{NiceTabular}{ccccccc}
	\begin{tabular}{cccccc}
		\toprule
		Scene &Method&PSNR$\uparrow$ &  SSIM$\uparrow$ &LPIPS$\downarrow$  \\
        \midrule
        \multirow{2}*{Doll}&Ours-joint&21.11& 0.634& 0.438\\
	  ~  & Ours& \textbf{24.22}& \textbf{0.787} &\textbf{0.228}\\
        \midrule
		\multirow{2}*{Bear}&Ours-joint& 22.99&0.678 & 0.517\\
	  ~  &Ours&  \textbf{25.12}& \textbf{0.788} &\textbf{0.291} \\
		\bottomrule
	% \end{NiceTabular}
	\end{tabular}
 }
	\label{tab:trainwide_dof}
\end{table}
% \begin{table*}[t]
% 	\centering
% 	\caption{{We conduct ablation experiments on our methods with the overexposed inputs. The best performance is in \textbf{boldface}.}} 
% 	\resizebox{1.0\linewidth}{!}
% 	{
%     \setlength{\tabcolsep}{3pt}
% 	\renewcommand\arraystretch{1.0}
% 	\begin{NiceTabular}{l|ccc|ccc|ccc}
% 		\toprule
% 		\multicolumn{1}{l}{\multirow{2}{*}[-0.5ex]{Input}} &
%   \multicolumn{3}{c}{Stadium} & \multicolumn{3}{c}{Tenhag}& \multicolumn{3}{c}{LAC}\\
% 		\cmidrule{2-10}
% 		~ &  PSNR$\uparrow$ &  SSIM$\uparrow$&LPIPS$\downarrow$ & PSNR$\uparrow$ & SSIM$\uparrow$ & LPIPS$\downarrow$ &PSNR$\uparrow$ &  SSIM$\uparrow$ & LPIPS$\downarrow$  \\
%         \midrule
% 		\midrule
%         % naive NeRF~\cite{mildenhall2020nerf}& 25.24 & 0.717 &0.422 &   25.72 & 0.722 &0.419& 25.48& 0.719 & 0.421\\
% 		Over Exposure &16.41 & 0.773&0.306& 14.38 & 0.598&0.411 & 16.81& 0.691& 0.407\\
%   Normal Exposure& \textbf{24.53} & \textbf{0.858} &  \textbf{0.178}& \textbf{24.53}&\textbf{0.737} & \textbf{0.254}& \textbf{22.65}&\textbf{0.748}&\textbf{0.315}\\
%   \bottomrule
%   \end{NiceTabular}
%  }
% 	\label{tab:exposure}
% \end{table*}

\begin{table}[!h]
	\centering
	\caption{Training and inference time of each NeRF method. These methods are tested on an Nvidia 3090 GPU.
 % , we only select the methods without pre-training models for deblurring each view. 
 % The best performance is in \textbf{boldface}.
 } 
	\resizebox{1.0\linewidth}{!}
 {
    \setlength{\tabcolsep}{3pt}
	\renewcommand\arraystretch{1.0}
	% \begin{NiceTabular}{ccccccc}
	\begin{tabular}{ccc}
		\toprule
		% \multicolumn{1}{c}{Framework Components} &\multicolumn{3}{c}{Framework Components} & \multicolumn{3}{c}{Total} \\
		% \cmidrule{0-5}
		Method & Training time (h)&Inference time (s)\\
        \midrule
        % wzj0503
		% \multirow{3}*{duck}& \checkmark & \times& 22.43& 0.67& 0.294\\
        Deblur-NeRF~\cite{ma2022deblur}& 18& 16 \\
        % \midrule
		DoF-NeRF~\cite{dofnerf}& {15}& 16\\
		DP-NeRF~\cite{Lee_2023_CVPR}& 22& 16\\
		\ourmethod  & 25& 30\\
		\bottomrule
	% \end{NiceTabular}
	\end{tabular}
 }
	\label{tab:time}
\end{table}

% \begin{figure}
%   \centering
%   \setlength{\abovecaptionskip}{3pt}
%     \setlength{\belowcaptionskip}{0pt}
%   \includegraphics[width=\linewidth]{fig/ablation_spatial.pdf}
%   \caption{Ablation study on our spatial warping operation. }
%   \label{fig:ablation_align}
% \end{figure}
\subsection{Applications}
\ourmethod can adjust the blur amount and the focal plane enabled by the disparity map and the defocus parameters. We demonstrate refocusing and split diopter in Fig.~\ref{fig:refocus} and Fig.~\ref{fig:split}. Refocusing shifts the focus point to a different depth plane, and split diopter keeps foreground and background objects in focus while the region between the objects is out of focus. The difference between split diopter and regular shallow DoF is split diopter does not have continuous DoF.

\subsection{Computational Cost}
We conduct the analysis of the computational cost for the total training time and the inference time for rendering one view in Table~\ref{tab:time}. Our NeRF model is not large, but we do have two MLPs for fusion. Therefore, the training cost and the inference time are more than those of other NeRF methods.

\subsection{Failure Case}
\begin{figure}[!htp]
  \centering
  \setlength{\abovecaptionskip}{3pt}
  \setlength{\belowcaptionskip}{-10pt}
  \includegraphics[width=\linewidth]{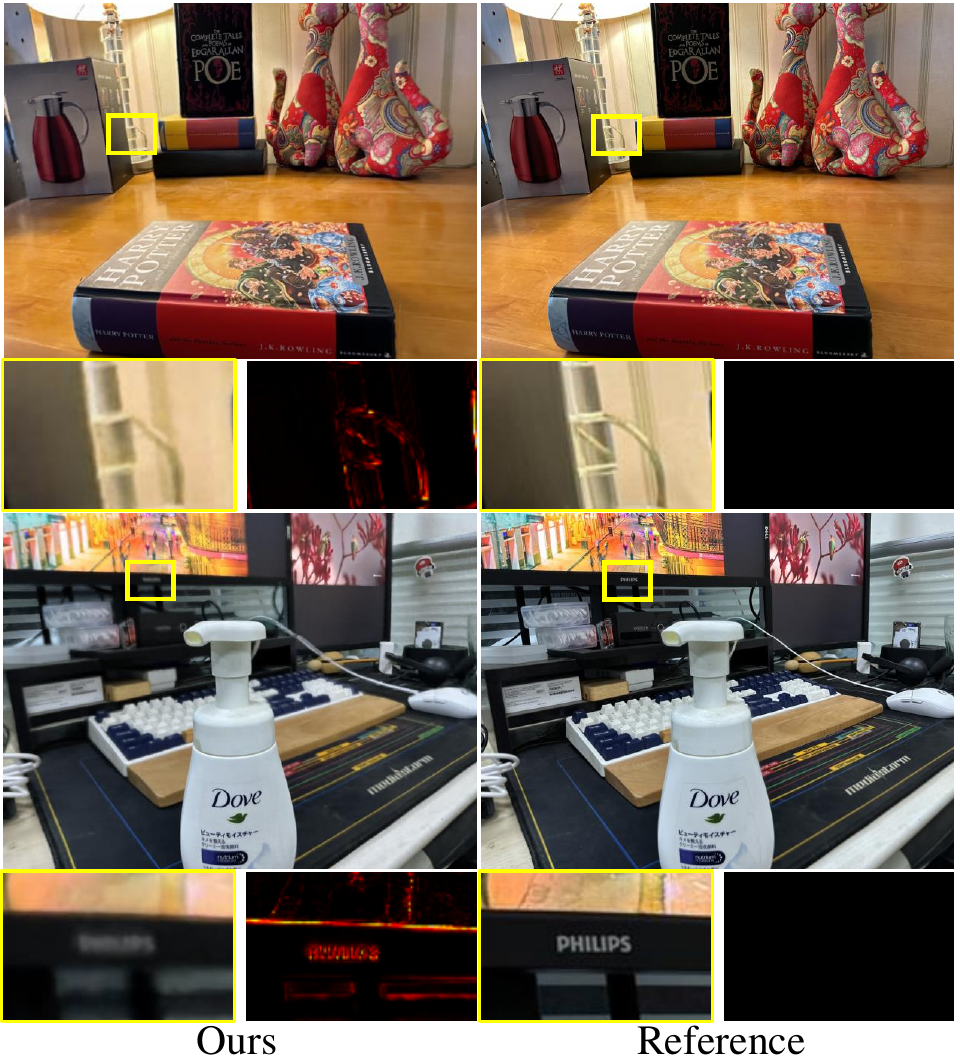}
  \caption{{Failure cases.} \ourmethod may have misalignment at edges because of defocus blur (row 1). The ultra-wide views may have low-quality regions on some details, so the corresponding parts of the main camera cannot be fixed (row 2). We compute the error map to show the difference, where darker regions indicate smaller errors.}
  \label{fig:failurecase}
\end{figure}

We show the failure cases in Fig.~\ref{fig:failurecase}. 
The optical flow network may suffer from artifacts due to severe defocus blur. Specifically, misalignments may occur at the edges because the flow computed from the defocused main camera is not accurate. Furthermore, the radiance field trained on ultra-wide camera rays may fail to render details because of the imaging quality. We plan to utilize additional cameras to jointly model the all-in-focus NeRF for future works.

\section{Discussion and Conclusion}
We present \ourmethod, a novel framework for all-in-focus novel view synthesis with the dual-camera system on smartphones.  
To deal with the situation where all input views have consistent blur and no sharp reference for some regions, the main/ultra-wide lens pair in smartphones is leveraged to combine the high quality of the main camera and the larger DoF from the ultra-wide camera. 
% we make use of the main/ultra-wide lens pair in smartphones to integrate both high-quality details from the main camera and large DoF from the ultra-wide one. 
We redefine the all-in-focus NeRF synthesis problem as an align-and-fuse solution. 
To align the two cameras for each view, spatial warping and color alignment are implemented. To fuse the contents from the aligned camera pair, we propose to learn the defocus parameters, then a blending mask is predicted from a blending weight field to fuse the main camera scene and the ultra-wide scene and generate all-in-focus novel views. We conduct experiments to show that our dual-camera all-in-focus NeRF is able to recover consistent defocus blur and produce satisfying results. 

Although this framework works well overall, some limitations remain to be addressed. The pre-trained optical flow network may produce unsatisfying details, and sometimes COLMAP may fail to estimate the camera poses. We plan to address these issues in future work.

\bibliographystyle{IEEEtran}
\bibliography{ref}

\vfill

\vfill

\ifCLASSOPTIONcaptionsoff
  \newpage
\fi

\end{document}